\documentclass{article}

\usepackage{microtype}
\usepackage{graphicx}
\usepackage{subfigure}
\usepackage{booktabs} 

\usepackage{hyperref}
\usepackage{multirow}
\usepackage{siunitx}

\usepackage[accepted]{icml2025}

\usepackage{amsmath}
\usepackage{amssymb}
\usepackage{mathtools}
\usepackage{amsthm}
\usepackage[inkscapelatex=false]{svg}

\usepackage[capitalize,noabbrev]{cleveref}

\theoremstyle{plain}

\theoremstyle{definition}

\theoremstyle{remark}

\newcommand{\floor}[1]{\mathrm{floor}\left(#1\right)}
\newcommand{\ceil}[1]{\mathrm{ceil}\left(#1\right)}

\usepackage[textsize=tiny]{todonotes}
\usepackage{textcomp}

\DeclareMathSizes{10}{9}{5}{4}

\icmltitlerunning{Accurate INT8 Training Through Dynamic Block-Level Fallback}

\begin{document}

\twocolumn[
\icmltitle{Accurate INT8 Training Through Dynamic Block-Level Fallback}



\icmlsetsymbol{equal}{*}

\begin{icmlauthorlist}
\icmlauthor{Pengle Zhang}{cs,zhili}
\icmlauthor{Jia Wei}{cs}
\icmlauthor{Jintao Zhang}{cs}
\icmlauthor{Jun Zhu}{cs}
\icmlauthor{Jianfei Chen}{cs}
\end{icmlauthorlist}
\icmlaffiliation{cs}{Department of Computer Science and Technology, Tsinghua University}
\icmlaffiliation{zhili}{Zhili College, Tsinghua University}

\icmlcorrespondingauthor{Jianfei Chen}{jianfeic@tsinghua.edu.cn}

\icmlkeywords{Machine Learning, ICML}

\vskip 0.3in
]



\printAffiliationsAndNotice{}  

\begin{abstract}
Transformer models have achieved remarkable success across various AI applications but face significant training costs. Low-bit training, such as INT8 training, can leverage computational units with higher throughput, and has already demonstrated its effectiveness on GPT2 models with block-level quantization. However, it struggles with modern Transformer variants incorporating GLU units. This is because those variants demonstrate complex distributions of activation outliers. 
To address the challenge, we propose Fallback Quantization, implementing mixed-precision GEMM that dynamically falls back 8-bit to 16-bit for activation blocks containing outliers. 
Experiments show that our approach is robustly competent in both fine-tuning and pretraining settings. Moreover, our method achieves a \textbf{1.57}$\times$ end-to-end training speedup on RTX4090 GPU.
\end{abstract}

\section{Introduction}

Recently, large-scale models based on the Transformer architecture have achieved remarkable success in natural language processing and computer vision. Models such as GPT-4~\cite{gpt4}, Llama~\cite{touvron2023llama}, and ViT~\cite{vit} have demonstrated state-of-the-art performance across various tasks. However, as both model parameter counts and data sizes continue to increase, training these large-scale models incurs significant computational costs. To accelerate the training process and reduce costs, researchers widely adopt low-precision numerical formats such as FP16 and BF16 for optimizing matrix multiplication operations \cite{micikevicius2017mixed}. Furthermore,
recent work has shown that fully quantized training (FQT) with even lower precision data formats, such as INT8 \cite{xi2024jetfire} and FP8 \cite{peng2023fp8}, is applied in Transformer training with promising results. FP8, first introduced in NVIDIA's Hopper \cite{nvidia2022hopper} architecture, offers a large dynamic range for training Transformer models, but its support is limited to a few specific hardware. INT8 has wider support across different platforms and is sometimes faster than FP8 operations.\footnote{For example, on RTX 4090, peak INT8 is 660Tops, which is 2x faster than the 330Tflops peak FP8 compute~\cite{nvidia2022ada}.} Despite the potential efficiency and hardware compatibility advantages, INT8 training is not yet as mature as FP8, since the narrow dynamic range makes it unsuitable to handle outliers in training. 
While early INT8 training works~\cite{banner2018scalable,zhu2020towards,chen2020statistical} adopted per-tensor quantization with applications on convolutional neural networks, there were some recent fine-grained quantization methods that succeeded in training transformers. Particularly, Switchback~\cite{switchback} quantized activation/gradient per-token and weight per-channel for training vision transformers. Jetfire~\cite{xi2024jetfire} proposed a more accurate per-block quantization method, where each $32\times 32$ block in activation/weight/gradient matrices had a separate scale. However, the small group size of Jetfire leads to large overhead in dequantization and accumulation, making the actual speedup unsatisfactory (only reaching 39\% peak flops on RTX 4090).
Moreover, Jetfire was only tested on GPT-2-style models~\cite{gpt2}, while recent architectures with GLU units and enlarged dataset~\cite{touvron2023llama} could be significantly harder to train.

\textbf{Motivation and our method.} In this work, we propose an accurate and efficient INT8 training method based on block-level mix-precision quantization. We observe that the activation outlier pattern in modern GLU-based architectures is sparse, and can be covered with a small fraction of square blocks. Therefore, we propose a \emph{dynamic block fallback quantization} method to allocate a higher bit-width to outliers when they are detected within specific quantization blocks. Compared with existing fine-grained quantization methods, this not only isolates the impact of outliers on the entire matrix but also improves the precision of the quantization block containing outliers themselves, which could carry critical information. 
Importantly, such mix-precision matrix multiplication (GEMM) can be implemented efficiently, similar to regular GEMMs, with minimal modifications.  
To further reduce memory consumption, we also integrate our training system with activation compression~\cite{chen2021actnn}, which stores fine-grain-quantized activations for backward calculation.

\textbf{Result and contribution.} First, on the algorithm side, our INT8 training method successfully solves a set of challenging finetuning and pretraining tasks on the strong Llama-3.1 and Qwen-2.5 models, with lossless accuracy and overlapping training curves with BF16 baselines. This is the first time that an INT8 training method can solve such hard problems. 
Second, on the kernel implementation side, our mix-precision GEMM kernel reached \textbf{425} TOPS on an RTX 4090, which is \textbf{2.58}$\times$ faster than BF16 and \textbf{1.65}$\times$ faster than Jetfire. Our framework achieves up to \textbf{1.57}$\times$ end-to-end speedup and \textbf{38}\% activation context memory reduction compared to BF16 training. 

\section{Related Work}

\textbf{Post Training Quantization} (PTQ)
~\cite{frantar2022gptq, yao2022zeroquant,xiao2023smoothquant, ashkboos2024quarot, zhang2025sageattention,zhang2024sageattention2,zhang2025sageattention3,zhang2025spargeattn,zhang2025sageattention2++,hu2025identifying} involves calibrating and converting pre-trained models to lower precision representations.
\textbf{Quantization Aware Training} (QAT)
~\cite{jacob2018quantization, dong2020hawq, liu2023llm-qat} integrates calibration and semi-quantization during the training process to enable models to adapt to quantization errors.  
Both methods focus on inference-time acceleration but retain high-precision formats during training.
Since they only focus on forward propagation with fixed model parameters, there exists more room for optimization.

\textbf{Fully Quantized Training} (FQT), on the other hand, employs low-precision data formats during training~\cite{banner2018scalable, cnn-fqt} and leverages corresponding hardware acceleration for training speedup. 
Current FQT approaches primarily utilize two data formats: FP8 and INT8. FP8, with its wider numeric range, is capable of handling training for many models~\cite{micikevicius2022fp8, perez2023training-fp8,peng2023fp8, fishman2024scaling} but is only supported on limited hardware platforms. 
INT8 computations are widely supported on most devices, but their narrow data range poses challenges for activation quantization. Previous works have employed fine-grained quantization~\cite{switchback,xi2024jetfire} to improve quantization accuracy. While block quantization has been successfully demonstrated on GPT2-scale models, its effectiveness has yet to be proven on recent transformers.

\textbf{Precision Fallback} is another strategy to reduce quantization errors by keeping certain data and computations at a higher bit-width.
Conventional mixed precision training maintains FP32 for non-matrix multiplication operations~\cite{micikevicius2017mixed, peng2023fp8}, but this approach is limited to the operator level.
Recent works, such as LLM.int8()~\cite{gpt3-int8, chen2024mixq}, preserve outliers in transformer activations at FP16 precision and utilize different TensorCores through channel shuffling. However, this approach is not suitable for dynamic training processes. \citet{int4-transformer} randomly preserves certain matrix computation elements at high precision to reduce errors, but fails to address the outliers in activation. Our block fallback method only performs fallback on activation blocks containing outliers, and is capable of utilizing only low-precision matrix multiplication units.

\section{Preliminary}

This section presents the preliminary quantization background and discusses INT8 training and its  challenges.

\subsection{Group Quantization}

\emph{Per-group quantization}~\cite{8bit-optimizer, frantar2022gptq} casts a $M\times N$ matrix $X$ to INT8 by 
partitioning it into quantization groups $\{G_{i,j}\}$ of size $M_g\times N_g$. Each group $G_{i,j}$ is transformed with a scale factor $a_{i,j}$ to scale its elements to the range $[-L, +L]$ ($L=127$), and then cast to the INT8 representation $\hat Q(G_{i,j})$ by rounding:
\[
\begin{aligned}
\hat Q(G_{i,j}) = \text{round}(G_{i,j}/a_{i,j}) \\
a_{i,j} = \max\{\text{abs}(G_{i,j})\}/L
\end{aligned}
\]
A low-precision representation can be \emph{dequantized} back to high-precision by multiplying the scale factor: $Q(G_{i,j}):=a_{ij} \hat Q(G_{i,j})\approx G_{i,j}$. For brevity and clarity, we also refer to it as block quantization.

For each value $x$ with $a=\floor{x}, b=\ceil{x}$, there are two possible rounding schemes: \emph{round-to-nearest} chooses the closest one from $x$, and \emph{stochastic rounding}~\cite{gupta2015deep} rounds $x$ to $b$ with probability $p=(x-a)/(b-a)$ and to $a$ with probability $1-p$, which is an unbiased approximation of the original full-precision value: $E[Q_{s}(x)] = x$.

\begin{figure}[t]
    \centering
    \subfigure[Group Quantization]{
        \includegraphics[width=0.515\linewidth]{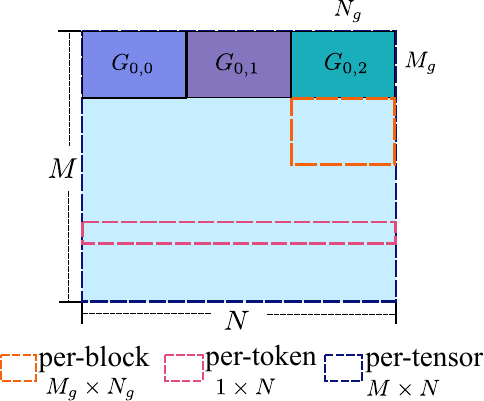}
        \label{fig:quantization}
        } \hfill
    \subfigure[TFLOPS/Group Sizes]{
        \raisebox{0.3cm}{\includegraphics[width=0.41\linewidth]{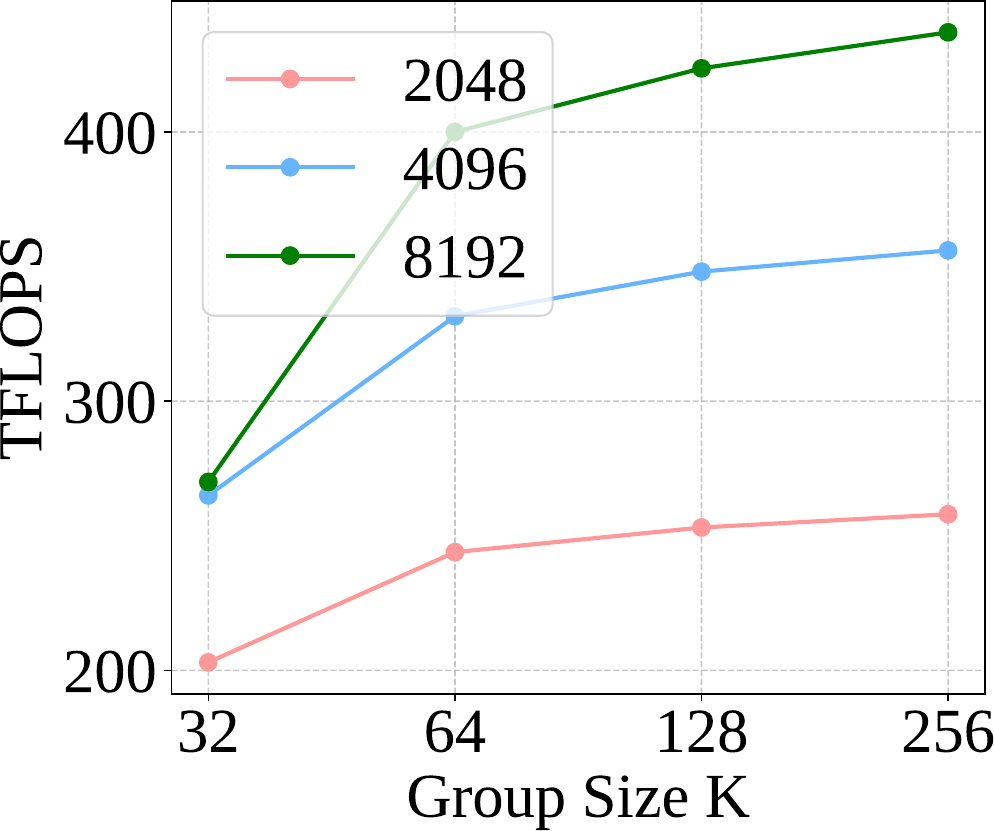}}
        
        \label{fig:groupsize-tflops}
    }
    \vspace{-.75em}
    \caption{(a) Different Quantization Methods. (b) Throughput performance with varying Group Size K on RTX4090, evaluated across different GEMM dimensions (2048, 4096, 8192).}
\end{figure}

Different quantization granularities correspond to different group sizes. Consider the case when $X$ is an activation matrix, where each row is a token, and each column is a channel. 
$M_g = M, N_g = N$ is per-tensor quantization, where the entire tensor has a single scale. 
$M_g = 1, N_g = N$ is per-token quantization, where each token has a single scale. Similarly, $M_g=M, N_g=1$ is per-channel quantization, and $M_g=\hat M, N_g = \hat N$  is per-block quantization with fixed group sizes $\hat M \times \hat N$.

\subsection{INT8 Training}

Low-precision training accelerates the computation of linear layers during training. Ignoring the bias term, the forward pass of a linear layer is one matrix multiplication (GEMM) $Y=XW^\top$, 
while the backward pass requires two GEMMs: $\nabla X = \nabla YW$ and $\nabla W = \nabla Y^\top X$ to compute the gradients of $X$ and $W$ respectively. Here, we denote weight/input/output gradients as $\nabla W/\nabla X/\nabla Y$. 
The training is accelerated by computing the above three GEMMs in both forward and backward passes in low-precision. For example, we can accelerate the GEMM $C=AB$ with quantization by computing $C\approx a^A a^B \hat Q(A)\hat Q(B)$, where $a^A, a^B$ are the scales for $A$ and $B$. The INT8 GEMM $\hat Q(A)\hat Q(B)$ can be computed 2x to 4x faster than FP16/BF16 GEMM.  

However, the accuracy of low-precision GEMM is problematic. Activations~\cite{bondarenko2021understanding} and gradients have many \emph{outlier} elements, which are orders of magnitude larger than other entries. Since the scale is determined by the maximum element in the group, the non-outlier elements will be very inaccurate. Previous works utilize fine-grained quantization to mitigate this problem. SwitchBack~\cite{switchback} adopts a per-token quantization for the input $X$ and a per-channel quantization for the weight $W$. However, it cannot handle the significant channel-wise outlier of the activation, so their experiments are mostly done on vision transformers rather than LLMs.

The recent Jetfire~\cite{xi2024jetfire} adopts a per-block quantization to all matrices with a group size $32\times 32$, and can handle both token- and channel-wise outliers. 
However, the small group size $32\times 32$ brings a large overhead of accumulation and dequantization. As shown in \cref{fig:groupsize-tflops}, on an RTX 4090 GPU, INT8 GEMM with $32\times 32$ group size can only reach 270 Tops, which is 38\% slower than the $128\times 128$ group size. Moreover, they only test GPT-2-style models, whether they can apply to modern architectures such as Llama is still questionable.

\section{Dynamic Block-Level Fallback}

One of the most important challenges in FQT is how to represent the high dynamic-range, rapidly changing activations accurately. 
This is particularly challenging for current GLU-based architectures, which have significantly larger outlier values. 
We propose a novel \emph{block fallback quantization} method to solve this problem. We start with analyzing the activation distributions of GLU-based networks.

\subsection{Outlier Pattern Analysis}
\label{sec:outlier-analysis}

We study the activation distribution in the latest Llama-3.1-8B and Qwen-2.5-7B, which are strong LLMs trained with trillions of tokens. Both models use GLU~\cite{glu}, which can be written as $y = \sigma(x_1) x_2$. Here, $\sigma(\cdot)$ is an activation function such as
SiLU~\cite{silu}, and $x_1, x_2$ are outputs of the previous linear layer. GLU computes the output by \emph{multiplying} two activations, which could amplify the magnitude, creating larger outliers~\cite{fishman2024scaling}, and making the quantization more difficult. 

In general, outliers make quantization challenging since the scale factor is determined based on the maximum element. The resolution (distance between adjacent quantization grid points) is coarser if the outlier is large. If the outlier is too large, it is likely that all  non-outlier entries are quantized to zero (\emph{underflow}), leading to significant information loss. On the other hand, the outlier entries themselves can carry much information, and might be very sensitive~\cite{bert-busters}, that a small quantization perturbation may cause large accuracy degradation. 
\begin{table}[th]
\caption{Maximum of outlier magnitudes at token, channel, tensor(others) levels on Llama-3.1-8B, Qwen-2.5-7B, OLMo-7B~\cite{OLMo}, GPT2-XL~\cite{gpt2} and Pythia-6.7B~\cite{biderman2023pythia} on WikiText. The latter two models do not have GLU. Outlier channels/tokens (top 5\% by L1-norm) and other outliers (extreme values outside outlier channels/tokens) are compared. }
\label{tab:outliers in modern transformers}
\vskip 0.15in
\begin{center}
\begin{small}
\begin{tabular}{lccc}
\toprule
Model & Token-wise & Channel-wise & Others \\
\midrule
\scriptsize Llama-3.1-8B  & \scriptsize ${605.77}$   &\scriptsize $123.54$  &\scriptsize $150.88$ \\
\scriptsize Qwen-2.5-7B & \scriptsize $6558.65$ &\scriptsize $622.83$  &\scriptsize ${574.80}$\\
\scriptsize Olmo-7B &\scriptsize ${375.95}$  &\scriptsize $265.12$& \scriptsize $270.10$ \\
\midrule
\scriptsize GPT2-XL &\scriptsize $56.28$ &\scriptsize $58.07$ & \scriptsize $30.14$ \\
\scriptsize Pythia-6.7B &\scriptsize $102.00$ &\scriptsize $100.05$ & \scriptsize $45.00$ \\

\bottomrule
\end{tabular}
\end{small}
\end{center}
\vskip -0.1in
\end{table}
\begin{figure}[htb]
    \centering
    \subfigure[Activation Distribution]{
        \includegraphics[width=0.42\linewidth]{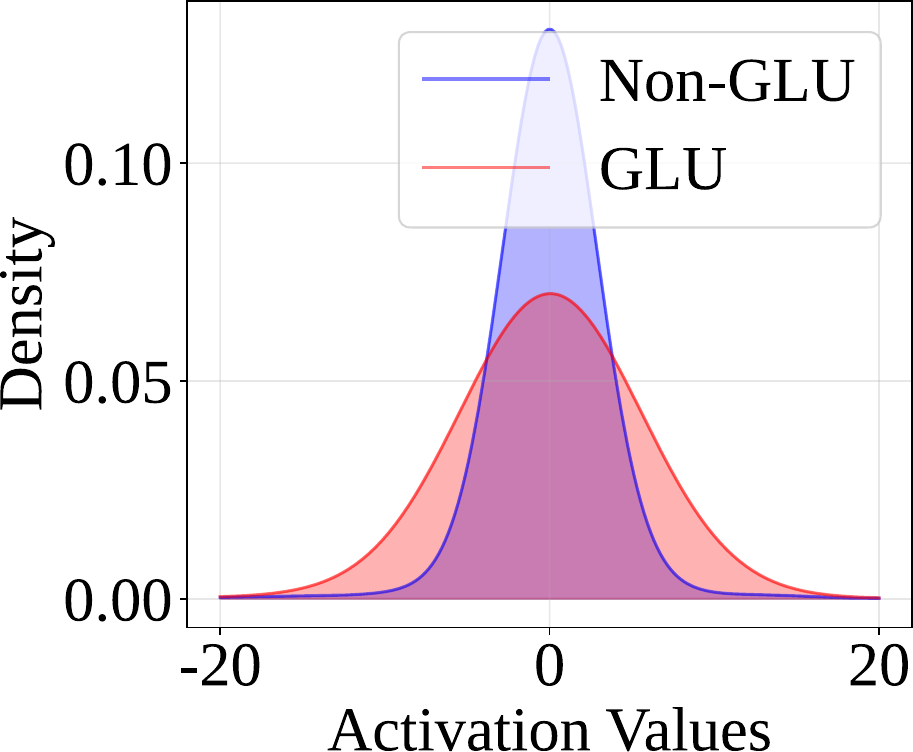}
        \label{fig:gpt2-activation-distribution}
    }
    \subfigure[Sorted Values]{
        \includegraphics[width=0.42\linewidth]{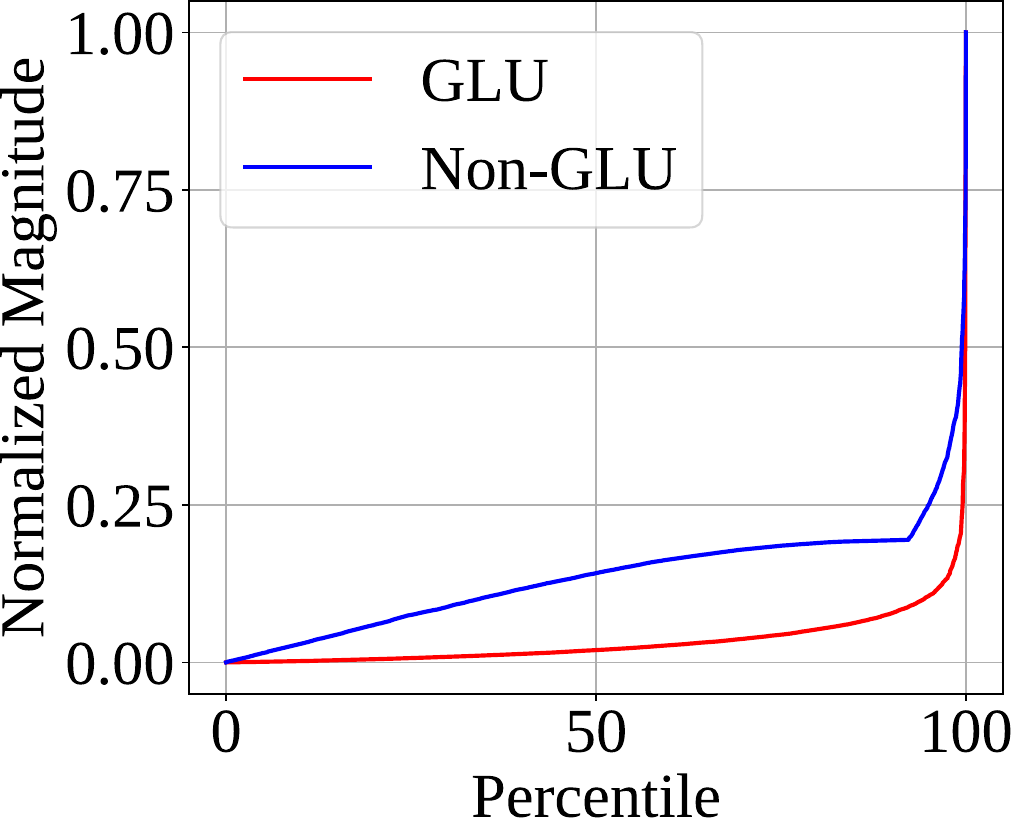}
        \label{fig:gpt2-sorted-values}
    }
    \subfigure[GLU Activation] {
        \includegraphics[width=0.9\linewidth]{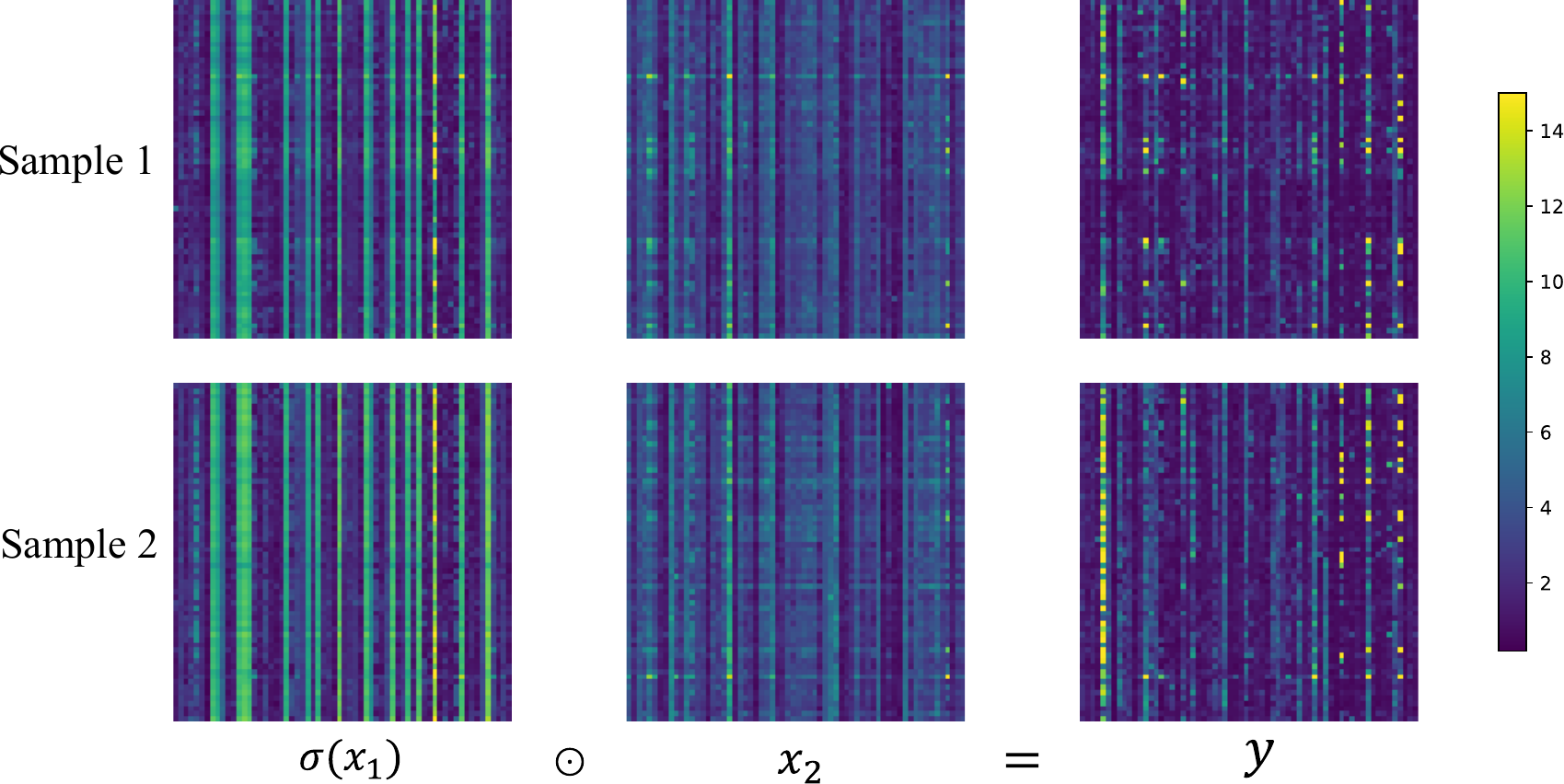}
        \label{fig:glu-activation}
    }
    \vspace{-.75em}
    \caption{
        GPT-2-Large models trained with identical hyperparameters (except for intermediate size), comparing GLU and non-GLU variants. Layer 20 analysis showing: (a) input distribution, (b) sorted magnitude distribution of normalized DownProj input. (c) GLU activation patterns in LLaMA-3.1-8B on WikiText. The magnitudes are truncated to 15 for better visual clarity.
    }
    
\end{figure}

There are some distinct outlier patterns compared to GPT-2-style models: \textbf{(P1) The outlier magnitude is significantly larger}. We analyze the maximum of outlier magnitudes on WikiText~\cite{wikitext}, in Table~\ref{tab:outliers in modern transformers}. 
While the outlier of GPT-2-style models does not exceed 127, the outlier of GLU-based models can be several hundreds or even several thousands due to the multiplicative nature of GLU. \textbf{(P2) Besides token and channel-wise outliers, there are also some \emph{occasional} outlier that appear randomly}. As shown in Table~\ref{tab:outliers in modern transformers} and Figure~\ref{fig:glu-activation}, even excluding outlier tokens and channels, there are still large outliers with magnitude on par with structured outliers. 
\textbf{(P3) the outlier pattern is sparse}.
As shown in Figure~\ref{fig:gpt2-sorted-values}, there are only a small fraction of elements that are much larger than others. The gating mechanism makes the activation sparser: $y$ is large only if both $\sigma(x_1)$ and $x_2$ are large. The sparsity holds even inside an outlier channel.

These patterns motivate the design of a new mixed-precision GEMM method. 
Specifically, due to sparse occasional outliers (\textbf{P2}) in any token/channel, both token-\cite{switchback} and channel-wise rescaling\cite{fishman2024scaling} are ineffective.
Although Jetfire's per-block quantization provides more flexible isolation of outliers from impacting an entire token/channel, blocks containing extremely large outliers (\textbf{P1}) still suffer from poor quantization resolution ($127/\max(\text{abs }G_{i,j})$). Most non-outlier values 
 in these blocks are inaccurate or quantized to zero (underflow).
This inspires the use of higher quantization resolution, i.e., higher precision.
Fortunately, since (\textbf{P3}) the outliers in GLU activations are highly sparse, we can improve quantization accuracy by retaining a small fraction of blocks containing outliers in higher precision, such as FP16/INT16. 

Based on the analysis above, we propose a \emph{block fallback quantization} method along with mixed-precision GEMM to realize this. At a high level, our method is a fine-grained \emph{mixed-precision} approach, where outliers and non-outliers have different numerical precision. The key to acceleration is that we need to design the mixed-precision algorithm in a hardware-friendly manner so we can preserve the accuracy while utilizing the fast TensorCores in hardware.

\subsection{GEMM with Block Quantization}
Before introducing our method, we first review Jetfire's
block-quantized GEMM. 
For the matrix $C$ of size $M\times N$ in GEMM $C = A\times B$, we partition it into $G^C_{i,j}$ of size $M_g \times N_g$, where calculation of each $G^C_{i,j}$ 
are independent tasks.
for $A$ and $B$ of sizes $M\times K$ and $K \times N$,
we partition them into tiles of shapes $M_g\times K_g$ and 
$K_g \times N_g$ to obtain $\{G_{i,k}^A\},\{G^B_{k,j}\}$ respectively. This breaks down the task into
accumulations of multiple sub-block matrix multiplications:
\[
G^C_{i,j} = \sum_{k=0}^{\lceil K/K_T\rceil-1}G^A_{i,k} G^B_{k,j}    
\]
Jetfire leverages this technique by quantizing each block of matrices $A$ and $B$ to INT8: $\hat Q(G^A_{i,k}), \hat Q(G^B_{k,j})$, and multiplies the dequantization coefficients {$a^A_{i,k} \times a^B_{k,j}$} during accumulation to achieve Block Quantization in low-precision GEMM:
\begin{align}
G^C_{i,j} &= \sum_{k=0}^{\lceil K/K_g\rceil-1} [\hat Q(G^{A}_{i,k})\hat Q(G^B_{k,j})]_{\text{INT}\times \text{INT}}(a^A_{i,k} \times a^B_{k,j})\label{eqn:int8} 
\end{align}

Here, $[\hat Q(G^{A}_{i,k})\hat Q( G^B_{k,j})]_{\text{INT}\times \text{INT}}$ takes INT8 inputs and outputs INT32, and result is then \emph{dequantized} and accumulated with an FP32 accumulator. 

\subsection{Block Fallback Quantization}

\begin{figure}[t]
    \centering
    \subfigure[Fallback Quantization]{
        \includegraphics[width=0.98\linewidth]{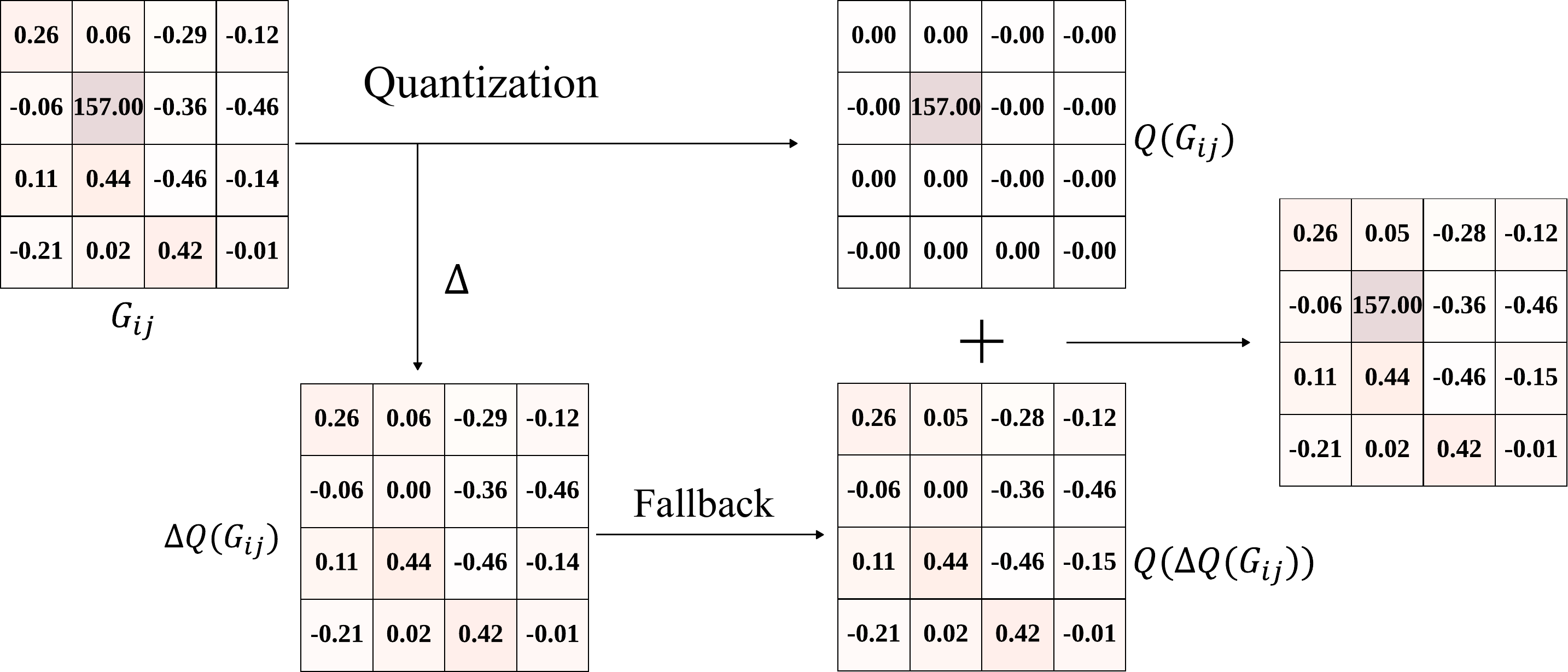}
        \label{fig:fallback-quantization}
    }
    \subfigure[RMSE/Bits]{
        \includegraphics[width=0.4\linewidth]{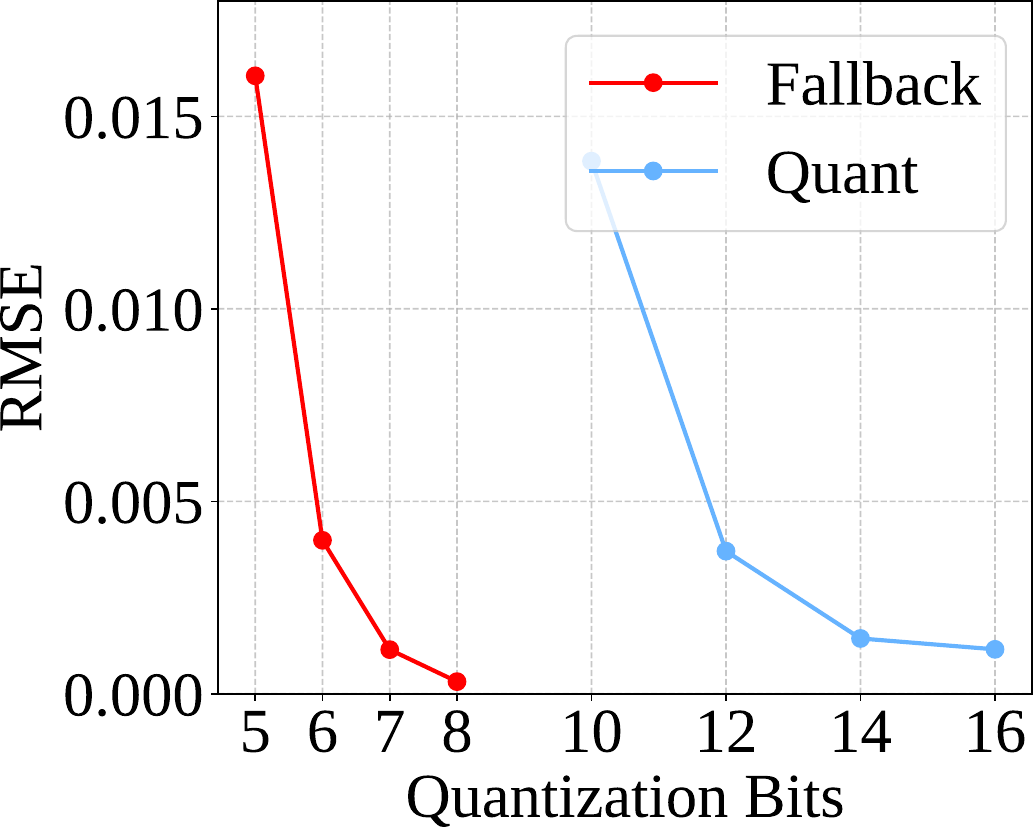}
        \label{fig:fallback-quant-rmse}
    }
    \subfigure[CosSim/FallbackRate]{
        \includegraphics[width=0.4\linewidth]{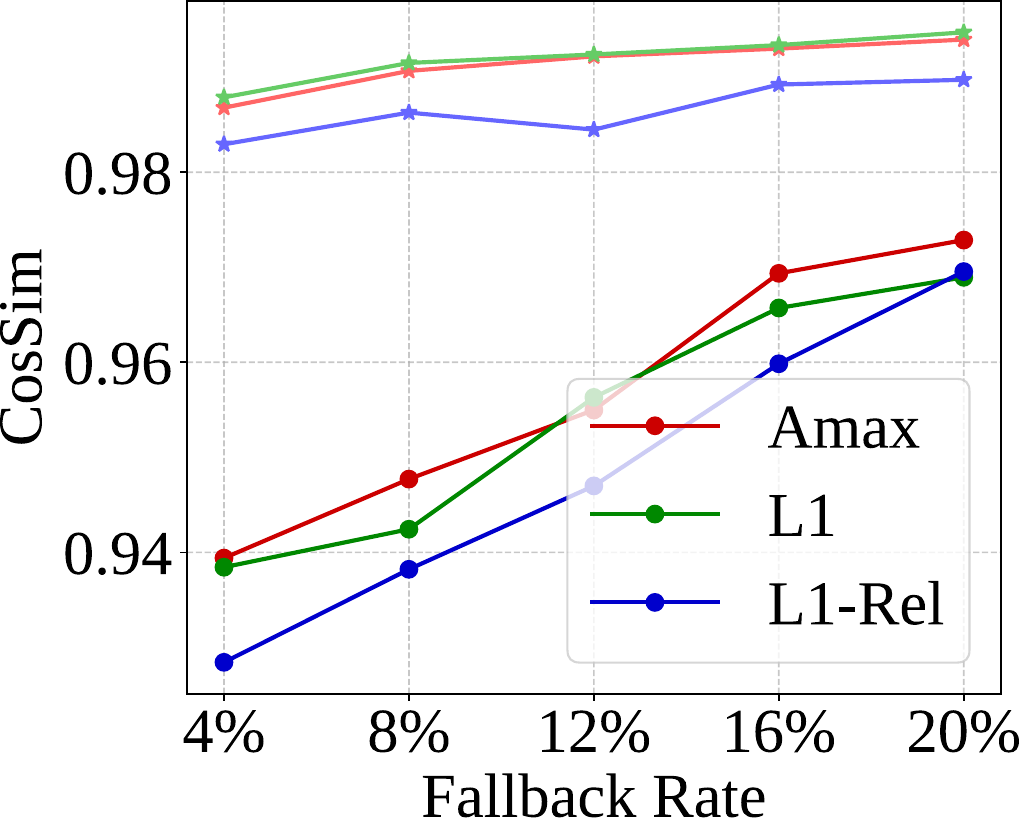}
        \label{fig:fallback-quant-cos-sim}
    }
    \vspace{-.75em}
    \caption{(a) Value underflow in naive INT8 block quantization for $G_{i,j}$ with outliers. Fallback Quantization captures the outlier in the first step and quantizes the remaining values in the second step. (b) RMSE comparison between Fallback and Double Bit block quantization on Qwen2.5-3B last layer activation. (c) Gradient CosSim across different Fallback criteria and rates.}
    
\end{figure}

We improve the accuracy of the per-block INT8 GEMM Eq.~(\ref{eqn:int8}) by adaptively detecting outlier blocks, and representing them in higher precision. Suppose the matrix $A$ has many outliers, and a specific quantization block $G^A_{i,k}$ is detected as an outlier block (will discuss in Sec.~\ref{sec:threshold}), we improve its precision by a two-step \emph{fallback quantization} procedure. 
In the first step, we quantize $G^A_{i,k}$ to obtain $Q(G^A_{i,k})$.
In the second step, we quantize the residual $\Delta Q(G^A_{i,k}) = G^A_{i,k} - Q(G^A_{i,k})$, resulting in a 16-bit representation $[Q(G^A_{i,k}), Q(\Delta Q(G^A_{i,k}))]$
of $G^A_{i,k}$,  as shown in Figure~\ref{fig:fallback-quantization}. Here, we call $Q(G^A_{i,k})$ the quantization block and $Q(\Delta Q(G^A_{i,k}))$ the fallback block of $G^A_{i,k}$. 

Conceptually, fallback quantization is similar to an INT16 representation. At first glance, fallback quantization might be less accurate since it only utilizes $(2^{8}-1)^2=65025$ quantization grid points, which is less than the $2^{16}-1=65535$ grid points provided by INT16. But in fact, fallback quantization is empirically more accurate than INT16, as shown in ~\ref{fig:fallback-quant-rmse}. This advantage stems from the sparsity of outliers within a quantization block (pattern \textbf{P3}) - these outliers can be filtered out in the first quantization step, thereby preserving the precision of other values in the second step (Figure~\ref{fig:fallback-quantization}).
Even in extreme cases, where a single exceptional outlier magnitude exceeds 20000~\cite{fishman2024scaling}, we can still maintain basic INT8 block quantization precision for the remaining values.

The main advantage of fallback quantization over simply keeping values in FP16/INT16 is that it can be easily integrated to the existing block-quantized GEMM kernel.
When applying block fallback quantization to matrix $A$, we have:
\[
{\tiny{G^C_{i,j}=\sum_{k=0}^{\lceil K/G_K\rceil-1} \left(Q(G^A_{i,k}) + u(i,k)Q(\Delta Q(G^A_{i,k}))\right) Q(G^B_{k,j})}}
\]
where $u(i,k) \in \{0, 1\}$ indicates whether the block $G^A_{i,k}$ undergoes fallback quantization. Compared to the original block-quantized GEMM, 
this GEMM only requires one $B$ block to compute with multiple $A$ blocks conditionally based on $u(i,k)$ 
as shown in Algorithm~\ref{alg:fallback-quantization-gemm}.
The performance impact is minimal: we only need to load the additional fallback block to compute an exact multiplication. The overhead is proportional to the ratio of fallback blocks. On the contrary, 
directly loading 16-bit data $G^A_{i,k}$ is complex and inefficient because of: (1) difficulty for on-chip memory management, as different precisions require different layouts. (2) $\hat Q(G^B_{k,j})$ has to be dequantized to $G^B_{k,j}$ on-chip to utilize 16-bit TensorCore, which is costly~\cite{lin2024qserve}.

\begin{algorithm}[htb]
   \caption{Fallback Quantization GEMM}
   \small
   \label{alg:fallback-quantization-gemm}
\begin{algorithmic}[1]
   \STATE {\bfseries Input:}
   \STATE Block Shape $[M_g, N_g, K_g]$
   \STATE Quantized $M\times K$ matrix $A$: $\{Q(G^A_{i,k})\}$
   \STATE Fallback Indicator $\{u(i,k)\}$
   \STATE Quantized $K \times N$ matrix $B$: $\{Q(G^B_{k,j})\}$
   \STATE {\bfseries Output:} Matrix $C$: $\{G^C_{i,j}\}$
   \FOR{$i=0$ {\bfseries to} $\lceil M / M_g\rceil$ - 1}
   \FOR{$j=0$ {\bfseries to} $\lceil N / N_g\rceil$ - 1}
   \STATE $G^C_{i,j} \leftarrow 0$
   \FOR{$k=0$ {\bfseries to} $\lceil K/ K_g\rceil-1$ \textcolor{red}{\textbf on-chip}}
   \STATE Load $\hat Q(G^A_{i,k}), \hat Q(G^B_{k,j}), a^A_{i,k}, a^B_{k,j}, u(i,k)$ to chip
   \STATE $G^C_{i,j} \mathrel{+}= [\hat Q(G^A_{i,k})\hat Q(G^B_{k,j})]_{\text{INT}\times\text{INT}}a^A_{i,k}a^B_{k,j}$
   \IF{$u(i,k) = 1$}
   \STATE Load $\hat Q(\Delta Q(G^A_{i,k})), \tilde a^A_{i,k}$ to chip
   \STATE $G^C_{i,j} \mathrel{+}= [\hat Q(\Delta Q(G^A_{i,k}))\hat Q(G^B_{k,j})]_{\text{INT}\times\text{INT}}\tilde a^A_{i,k}a^{B}_{k,j}$
   \ENDIF
   \ENDFOR
   \STATE Save $G^C_{i,j}$ to Memory $C$.
   \ENDFOR
   \ENDFOR
   \STATE {\bfseries Output} $C$
\end{algorithmic}
\end{algorithm}

\subsection{Threshold for Dynamic Fallback}
\label{sec:threshold}

As discussed in Section~\ref{sec:outlier-analysis},
to maintain accurate outlier and non-outlier values,
quantization resolution
should be fine-grained through representing outliers 
in higher precision.
Following this principle, we can determine block fallback decisions ($u(i,k)$)
based on the AbsMax ($\max(\text{abs }G^A_{i,k})$),
where the TopK AbsMax $A$ quantization blocks undergo fallback.
Another straightforward approach is to select blocks based 
on their overall block quantization error, which can be measured 
using either absolute (L1: $L_1^Q(G^A_{i,k})=\sum\text{abs}(G^A_{i,k}-Q(G^A_{i,k}))$) or relative metrics (L1-Rel: $L_1^Q(G^A_{i,k})/\sum\text{abs }G^A_{i,k}$).

Experimental results on Qwen2.5 models in Figure~\ref{fig:fallback-quant-cos-sim} show that, 
AbsMax and L1 error metrics demonstrate similar
effectiveness, while related error-based selection 
shows inferior compensation for gradient errors.
Given that AbsMax values are readily available 
from the first quantization step, we adopt
AbsMax as our fallback selection threshold 
$u(i,k) = [\max(\text{abs }G^A_{i,k}) > \theta]$.
Based on this, we examined the distribution of
fallback blocks with overall fallback rate 20\% in Qwen2.5-3B, illustrated in Figure~\ref{fig:fallback-sample},
corroborates our earlier findings: 
dynamic fallback is necessary 
for blocks containing occasional outliers,
while preserving per-channel outliers.

{Directly selecting TopK AbsMax as the Threshold requires tensor-level reduction which leads to performance issues. Instead, we use the Delay Threshold method to maintain the fallback rate within a range $[r_{min}, r_{max}]$ with an adjustment factor $\alpha$. This process is described in Appendix~\ref{appendix:delay-threshold}.}

\subsection{Kernel Implementation for better Acceleration}

\begin{figure}[t]
    \centering
    \subfigure[Fallback Block]{
        \includegraphics[width=0.35\linewidth]{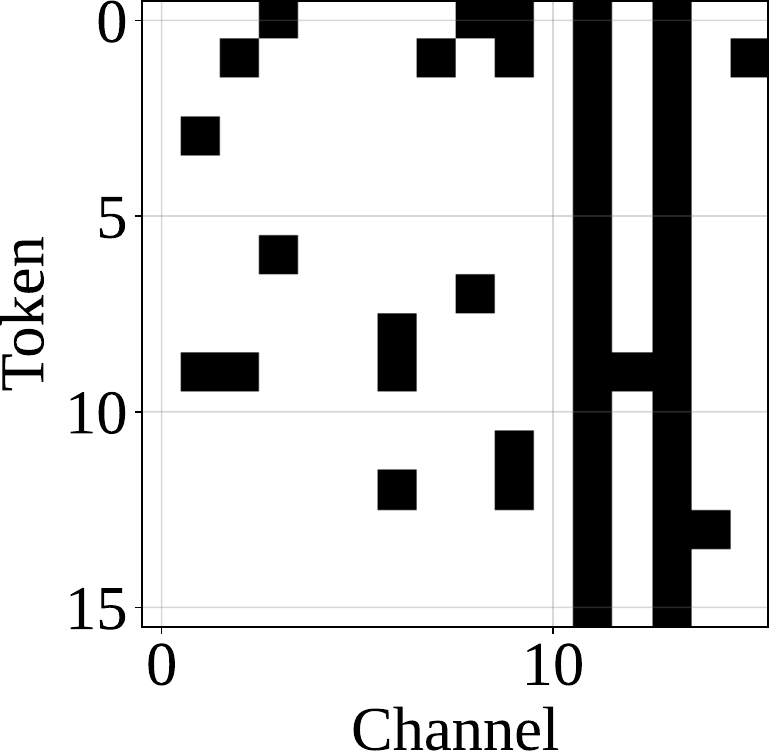}
        \label{fig:fallback-sample}
    }
    \subfigure[PPL/BlockSize]{
        \includegraphics[width=0.42\linewidth]{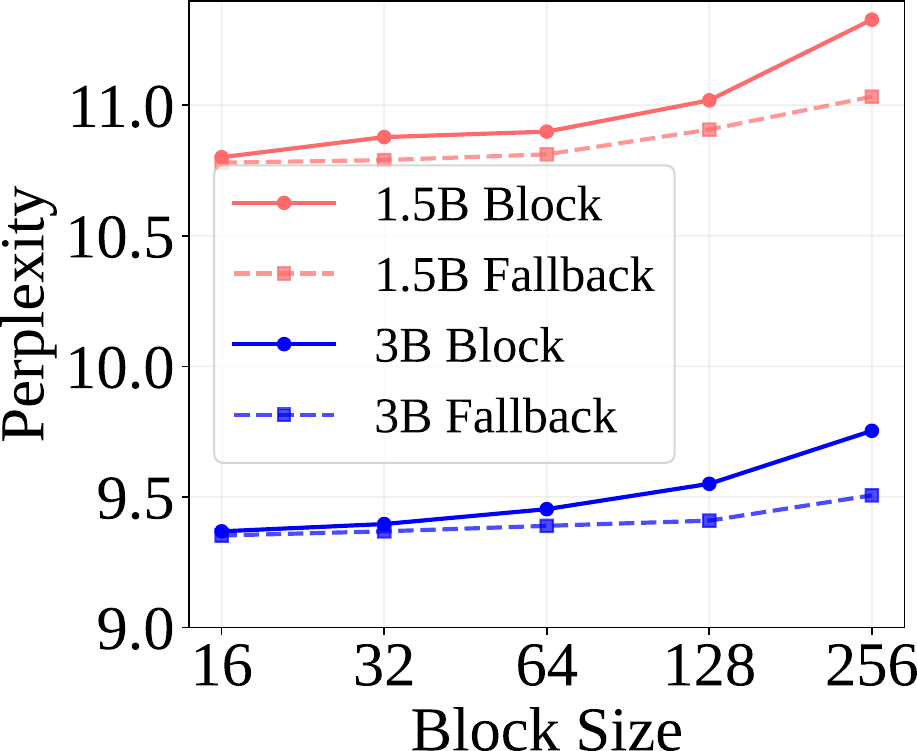}
        \label{fig:fallback-quant-ppl}
    }
    \vspace{-.75em}
    \caption{(a) Fallback block distribution in the last layer Down-Proj Input of Qwen-2.5-3B. (b) Perplexity comparison of naive and Fallback (20\% AbsMax) INT8 quantization across different block sizes on Qwen2.5 models.}
\end{figure}

The fallback mechanism enhances the accuracy of block activation
quantization, enabling a large block size of 128 with fallback to achieve nearly the same precision as a small block size of 32, as shown in Figure~\ref{fig:fallback-quant-ppl}.
Since block size 32 is the performance bottleneck in Jetfire's GEMM, we adopted a block size of 128 for better acceleration as illustrated in Figure~\ref{fig:groupsize-tflops}.

In Algorithm~\ref{alg:fallback-quantization-gemm}, we need to set specific numbers for the block size of GEMM $M_g\times N_g \times K_g$. However, in real GEMM implementation, we need to adjust the tile size~\cite{cutlass} of matrices for different  GPUs. Note that this adjustment does not conflict with the specific numbers of  $M_g\times N_g \times K_g$. This is because the tile size is normally smaller than the block size $128 \times 128 \times 128$, leading to the feasibility of implementing GEMM in the range of the quantization blocks. We leave the details in Appendix~\ref{appendix:different-block-sizes}.

\section{Training System Design}

Following our discussion of the core fallback mechanism, we now detail the quantization of Linear and Non-Linear layers, and the implementation of the training framework.

\subsection{Linear Layer}

\begin{figure}[t]
    \centering
    \subfigure[CosSim/Bits]{
        \includegraphics[width=0.42\linewidth]{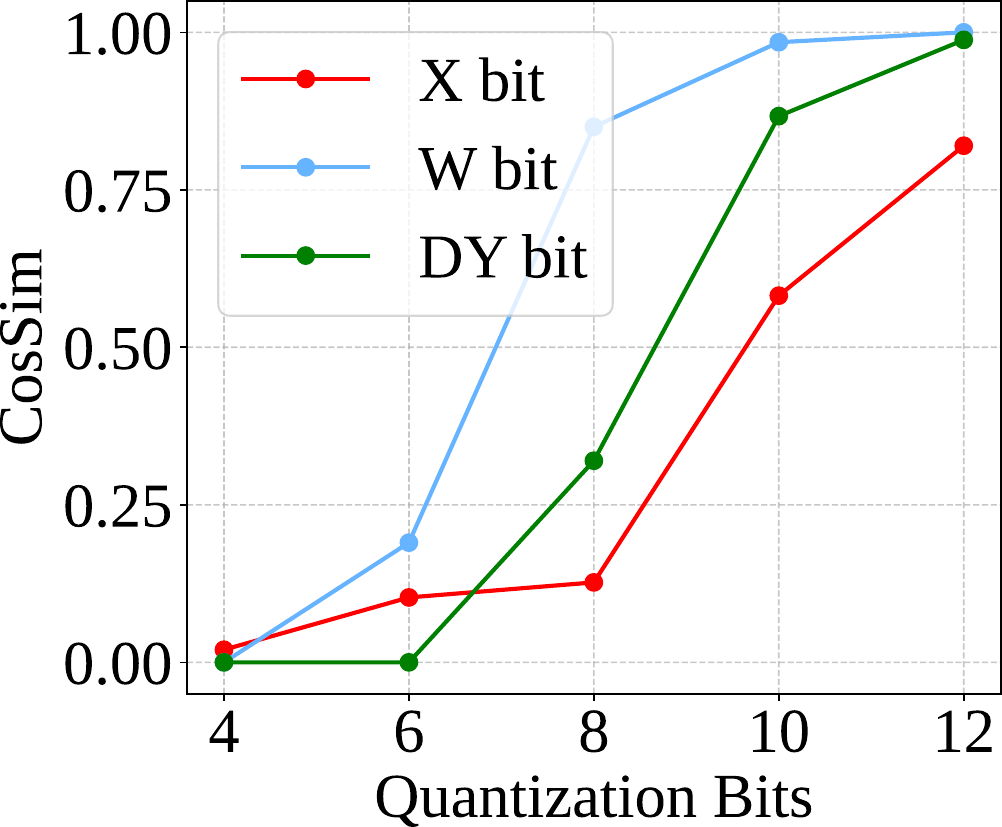}
        \label{fig:quant-bits-cossim}
    }
    \subfigure[CosSim/FallbackRate]{
        \includegraphics[width=0.42\linewidth]{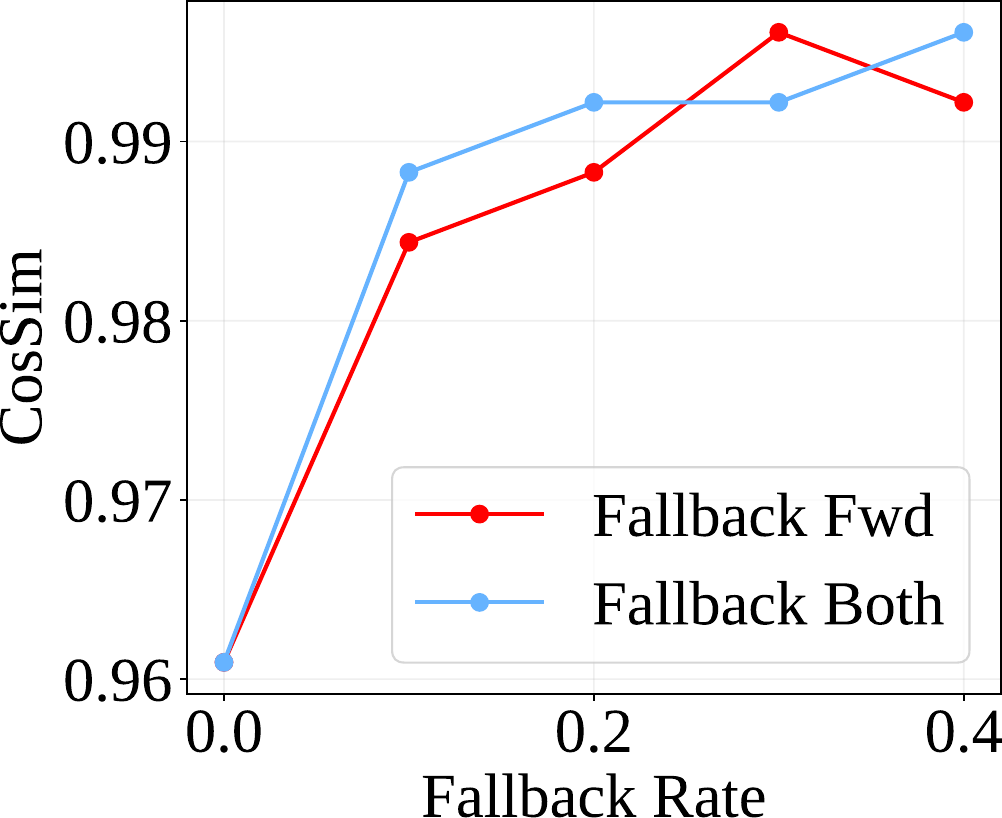}
        \label{fig:fallback-x-cossim}
    }
    \vspace{-.75em}
    \caption{Gradient CosSim (a) for different per-tensor bit quantization of $X, W, \nabla Y$ (b) when applying fallback to X in forward only or in both forward and backward passes on Qwen2.5-1.5B.}
\end{figure}

While both $X$ and $\nabla Y$ exhibit large outliers~\cite{int4-transformer}, we only adopt fallback quantization for $X$.
This design choice is based on several considerations.
First, the quantization error of $\nabla Y$ can be effectively mitigated through stochastic rounding, which maintains consistency with SGD theoretical requirements~\cite{chen2020statistical}.
As shown in Figure~\ref{fig:quant-bits-cossim}, when using stochastic rounding for $\nabla Y$, the quantization error of 
$X$ contributes the largest portion of gradient errors.
Additionally, since $\nabla Y$ participates in two GEMMs during the backward pass, using standard block GEMM
is more effective at improving throughput. Therefore, we do not apply fallback quantization to $\nabla Y$.

The $X$ participates in GEMM twice during the forward and backward passes: $Y=XW^T$ and $\nabla W=\nabla Y^T X$. We have to preserve activation context higher than 8-bit if we apply fallback in both GEMMs. 
However, 
since $\nabla W = E[\nabla Y^T]E[X]$, we can 
still utilize stochastic rounding for $X$ 
to simplify memory management and employ standard block GEMM for better efficiency.
Our experiments confirm that these two approaches show no significant difference in accuracy as illustrated in Figure~\ref{fig:fallback-x-cossim}.

\subsection{Non-Linear Layer}

For Non-Linear layers such as Normalization and Activation functions, we can flexibly compress/decompress their activations since they are not constrained by TensorCore data format requirements. Jetfire adopts INT8 data flow for these layers, with inputs and outputs quantized using $32\times 32$ block size to optimize memory footprint and improve throughput.

\begin{figure}[t]
    \centering
    \subfigure[Sensitivity]{
        \includegraphics[width=0.40\linewidth]{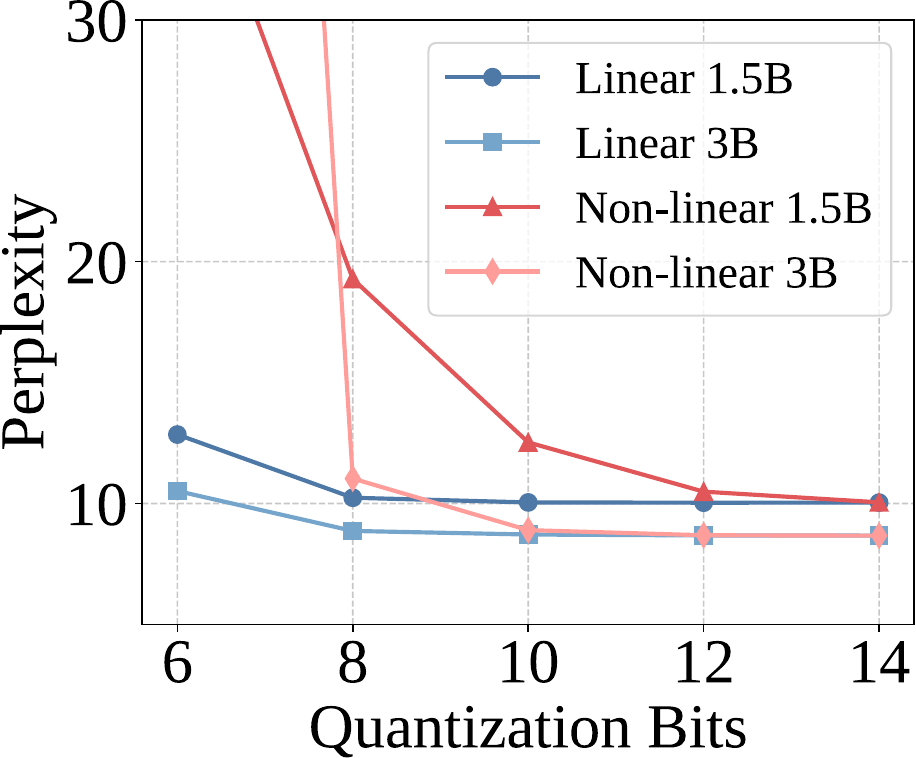}
        \label{fig:non-linear-ppl}
    }
    \subfigure[Time/ModelSize]{
        \includegraphics[width=0.47\linewidth]{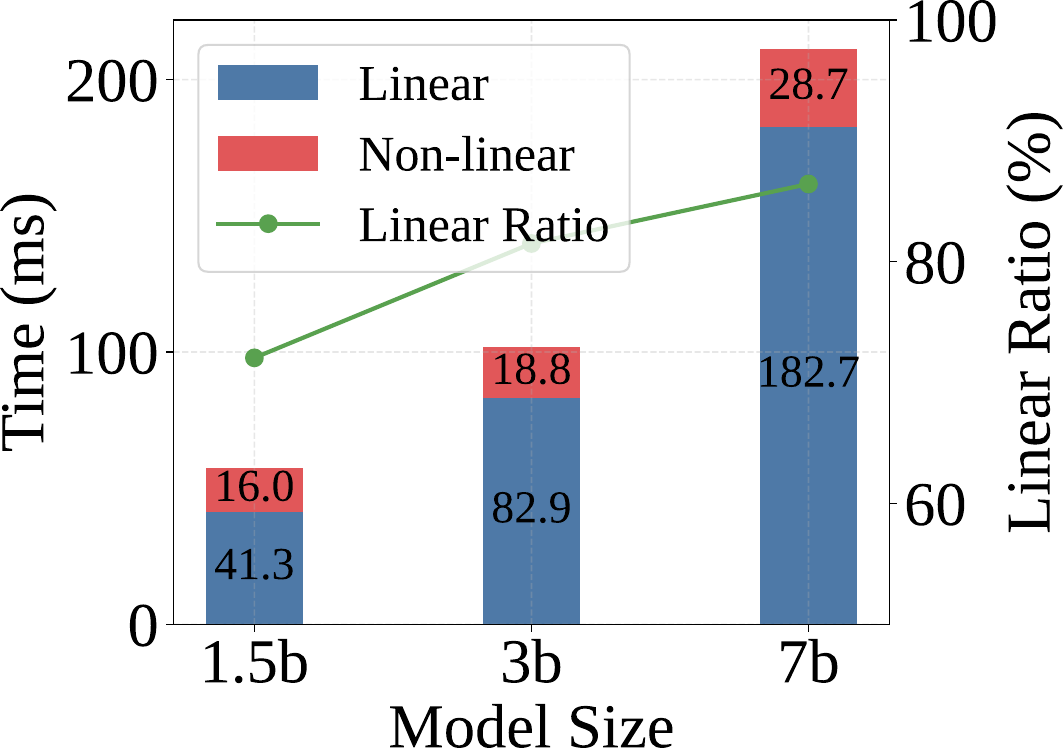}
        \label{fig:non-linear-ratio}
    }
    \vspace{-.75em}
    \caption{(a) Impact of different bit-width 
 block quantization on Linear and Non-Linear inputs on the PPL of Qwen2.5 1.5B and 3B models. (b) The proportion of computation time spent on Linear layers across different sizes of Qwen2.5 models in forward-pass.}
\end{figure}

However, Non-Linear layers are particularly sensitive to quantization errors (Figure~\ref{fig:non-linear-ppl}),
as they cannot mitigate errors through $K$ dimension accumulation like Linear layers.
Moreover, their computational complexity scales linearly with model size in contrast to the quadratic scaling of 
Linear layers, resulting in marginal optimization returns in large models as presented in Figure~\ref{fig:non-linear-ratio}.

\begin{figure}[t]
    \centering
    \subfigure[Non-linear/Bits]{
        \includegraphics[width=0.43\linewidth]{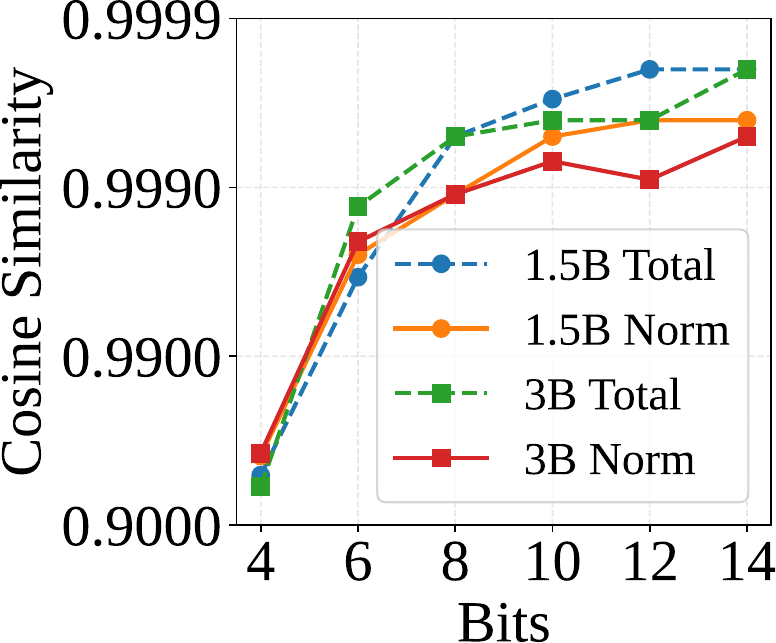}
        \label{fig:non-linear-bit}
    }
    \subfigure[Pretrain] {
    \includegraphics[width=0.42\linewidth]{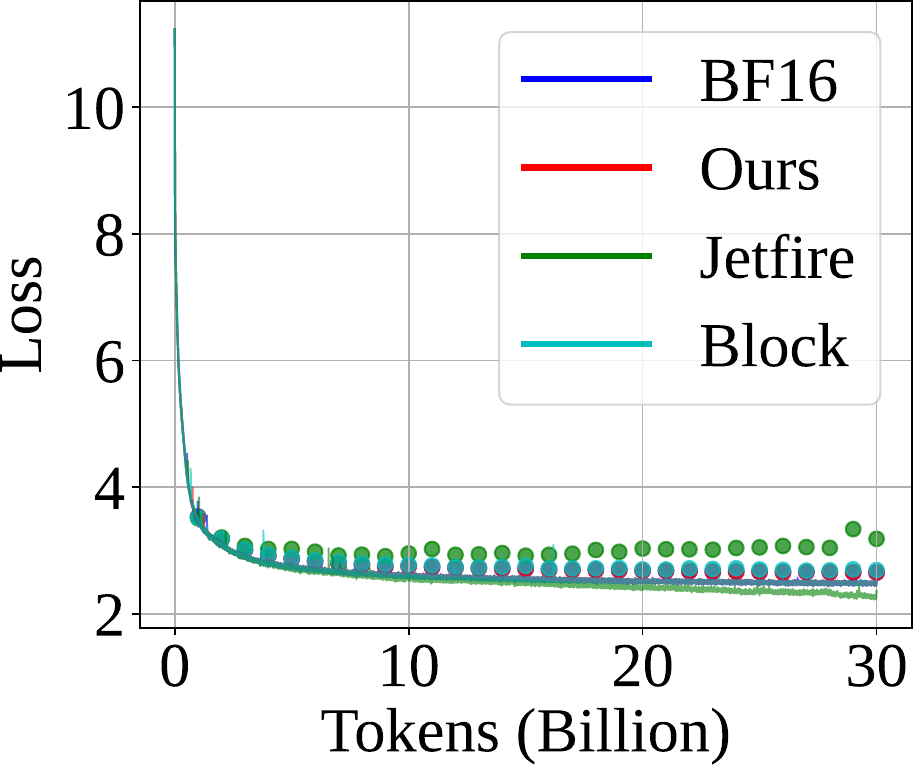}
    \label{fig:pretrain-loss}
    }
    \\ \vspace{-.75em}
    \subfigure[Attention]{
        \includegraphics[width=0.42\linewidth]{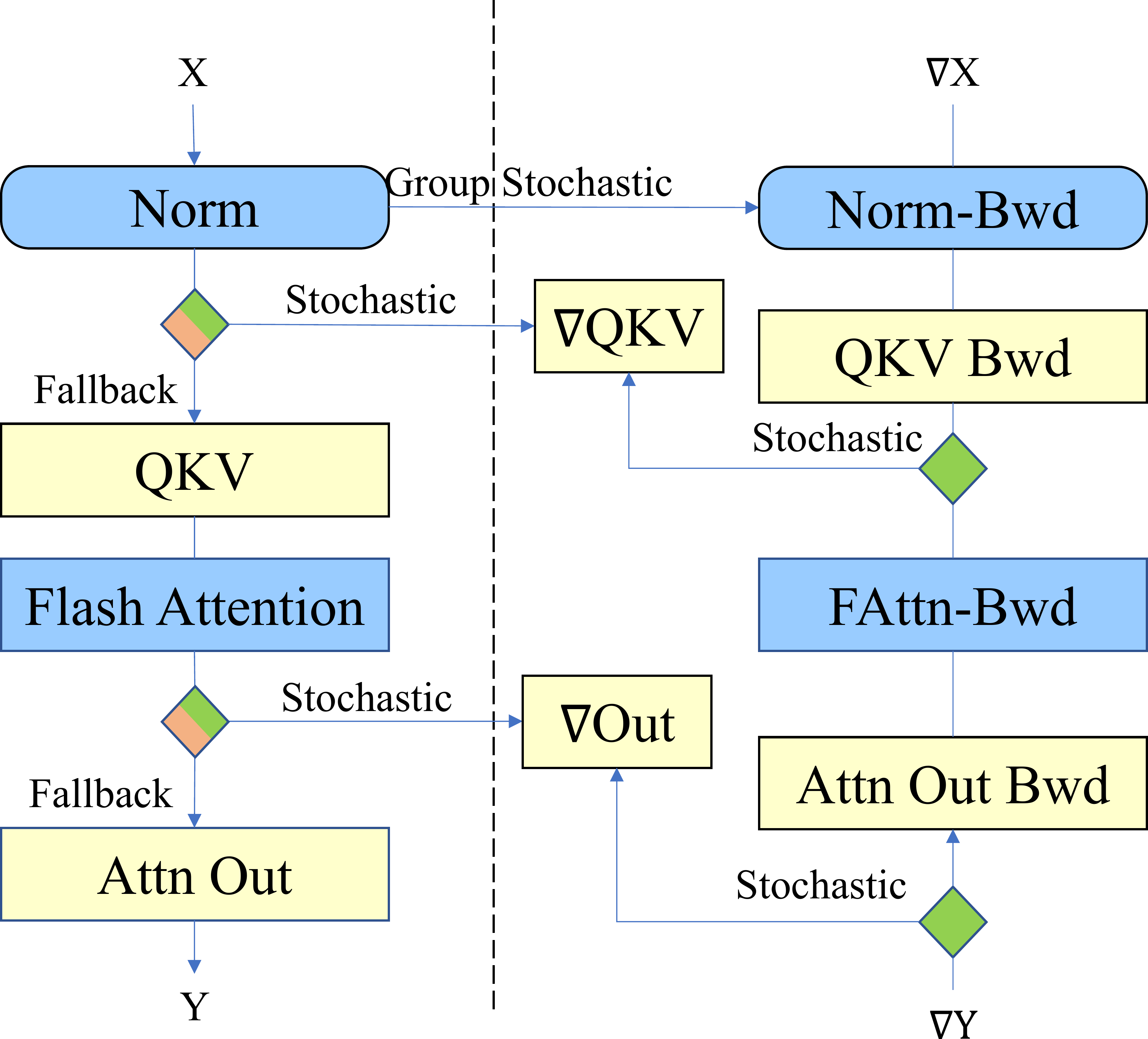}
        \label{fig:attention-mf}
    }
    \subfigure[MLP]{
        \includegraphics[width=0.42\linewidth]{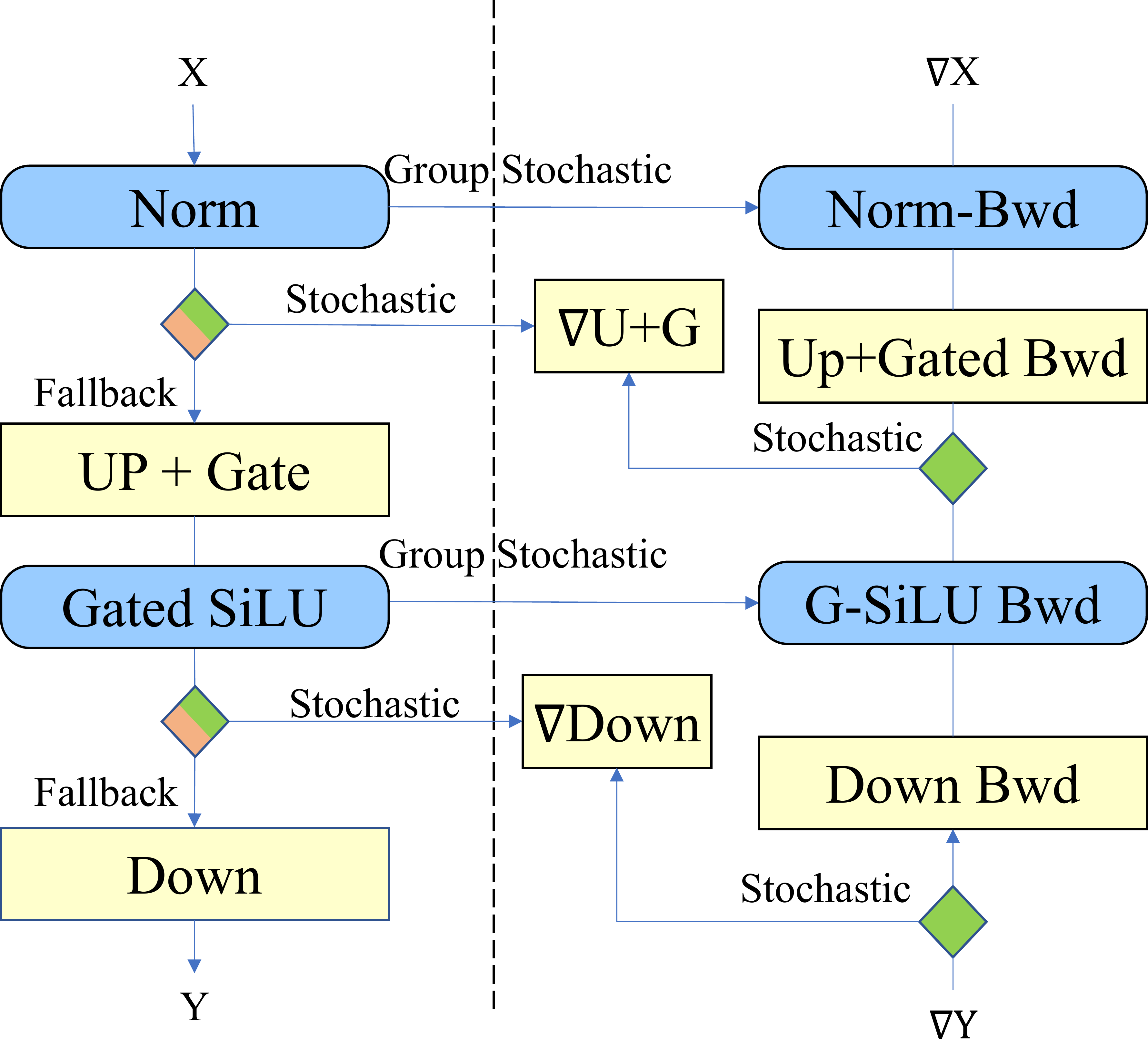}
        \label{fig:mlp-mf}
    }
    \vspace{-.75em}
    \caption{(a) Cosine similarity analysis of model gradient and norm weight gradient under different bits of group quantization on Qwen2.5. (b) Pretrain and Validation Loss (dots). (c), (d) Training framework data-flow overview.}
\end{figure}

Memory consumption is another critical aspect of Non-Linear layers, as they produce
activation contexts comparable to Linear layers.
To maintain their accuracy while reducing activation contexts,
we utilize $1\times 128$ per-group INT quantization to enable per-token processing in kernel fusion.
We evaluated model gradient errors under various compression bits and found that 10-bit integer quantization achieves near-lossless results as shown in \cref{fig:non-linear-bit} while reducing memory usage to 5/8 of BF16.

\subsection{Training Framework Implementation}

Building upon our previous discussion, we present our training framework, which shares similarities with Jetfire. The framework incorporates both Linear and Non-Linear operations with specific optimizations for each component.

For linear layers, we employ fallback quantization GEMM in the forward pass while maintaining the block stochastic quantization of input $X$ as the activation context. In the backward pass, $\nabla Y$ undergoes stochastic quantization followed by two block-quantized GEMM operations. For Normalization and Activation functions, we preserve their input and output in BF16 precision while quantizing their activation contexts to INT10 with $1\times 128$ group size. These contexts are dequantized during the backward pass for gradient computation.  
For Dot Production Attention, as it fuses Linear and Non-Linear operators into the Flash Attention~\cite{dao2023flashattention} kernel, its quantization requires more detailed analysis. Therefore, we maintain it in BF16 precision.

To further improve efficiency, we fuse dynamic fallback quantization into a quantization kernel and fuse the Non-Linear operations: the forward pass with group quantization and the backward pass with dequantization.
The complete training framework is illustrated in \cref{fig:attention-mf,fig:mlp-mf}, showing the data flow through Attention and MLP modules.

\section{Experiment}
\label{section:exp}

\begin{table*}[!t] 
\centering
\caption{Finetune results. CAL-FLOPS represents the matrix multiplication throughput per microstep on RTX4090; only computation time is measured. ACT-MEM denotes the GPU memory consumption of activation contexts.}
\sisetup{
    round-mode = places, 
    round-precision = 3 
}
\label{tab:finetune}
\scriptsize
\begin{tabular}{
    cc 
    S[table-format=1.2]
    S[table-format=1.2]
    S[table-format=1.2]
    S[table-format=1.2]
    cc
}

\toprule
\footnotesize{Model} 
& \footnotesize{Method} 
& \footnotesize{GSM8K}\tiny{(Acc)} 
& \footnotesize{DROP}\tiny{(F1)}
& \footnotesize{MMLU}\tiny{(Acc)} 
& \footnotesize{HELLASWAG}\tiny{(Acc)} 
& \footnotesize{CAL-FLOPS\tiny(T)} & \footnotesize{ACT-MEM\tiny(GB)}\\
\midrule
\multirow{4}{*}{\small{Qwen2.5-1.5B}} & BF16 
&0.521607278 & 0.644156754 &  0.560176613 & 0.907986457 & 112.37 & 3.92 \\
\cmidrule{2-8}
& Block 
& 0.005307051 &0.651484934&0.561031192 & 0.904700259 & 158.07 & 3.20\\
& Jetfire
& 0.441243366 & 0.63566513018532 & 0.54465175900869 & 0.28888667596096 & - & 2.08 \\
\cmidrule{2-8}
& Ours 
& 0.510993177 & 0.636414163 & 0.552841476 &  0.900517825 & 154.81{\tiny(1.38x)} & 2.39{\tiny(61\%)} \\
\midrule
\multirow{4}{*}{\small{Qwen2.5-3B}} & BF16 
& 0.589840788 & 0.655050891 & 0.610739211  & 0.928400717 & 125.23 & 4.91 \\
\cmidrule{2-8}
& Block 
& 0.584533738 & 0.672323135 & 0.60653753 & 0.927803226 & 197.87 & 4.00 \\
& Jetfire 
& 0.60879454131918 & 0.69203290999043 & 0.60205098988748 & 0.92172873929496 & -  & 2.61 \\
\cmidrule{2-8}
& Ours 
& 0.583775588 & 0.670923463 & 0.600555476 & 0.929396535 & 186.87{\tiny(1.49x)} & 2.99{\tiny(61\%)} \\
\midrule
\multirow{4}{*}{\small{Llama-3.2-1B}} & BF16 
& 0.26535254 & 0.520773407 & 0.420025637 & 0.816968731 & 118.15 & 4.07 \\
\cmidrule{2-8}
& Block 
& 0.26459439 & 0.526532737 & 0.389972938 & 0.827623979 & 168.44 & 3.30 \\
& Jetfire 
& 0.24184988627748 & 0.51602534747884& 0.40222190571144& 0.82523401712806 & - & 2.21 \\
\cmidrule{2-8}
& Ours 
& 0.255496588 & 0.525712389 & 0.421806011 & 0.831408086 & 164.78{\tiny(1.39x)} & 2.52{\tiny(62\%)}\\
\midrule
\multirow{4}{*}{\small{Llama-3.1-8B}} & BF16 
& 0.474601971 & 0.596569942 & 0.519584105 & 0.910476001 & 135.31 & 3.74 \\
\cmidrule{2-8}
& Block 
& 0.475360121 & 0.60698269 & 0.525423729 & 0.904799841 & 216.36 & 3.03 \\
& Jetfire 
& 0.47915087187263 & 0.61768347722757 & 0.52528129896026 & 0.91017725552679 & - & 2.05 \\ 
\cmidrule{2-8}
& Ours 
& 0.492797574 & 0.588826391 & 0.523785786 & 0.912766381 & 212.01{\tiny(\textbf{1.57x})} & 2.32{\tiny(62\%)} \\
\bottomrule

\end{tabular}
\end{table*}

In this section, we evaluate our method through both finetuning and pretraining experiments, demonstrating its effectiveness and performance gains.

\textbf{Setup.} In all experiments, we set the fallback range to [0.1, 0.3] and use an adjustment factor of $\alpha=1.3$
, with 10-bit quantization for Context in Non-Linear Operations. Detailed setting is in Appendix~\ref{appendix:exp-set}. 

\subsection{Fine-tuning}

We evaluate the effectiveness of different methods using Qwen2.5 1.5B and 3B, Llama-3.2-1B, and Llama-3.1-8B on GSM8K~\cite{cobbe2021gsm8k}, DROP~\cite{Dua2019DROP}, MMLU~\cite{MMLU}, and HELLASWAG~\cite{zellers2019hellaswag} datasets.
All methods use identical training configurations.
Our baselines include the original BF16 training, block-quantized GEMM (Block) with quantization only in Linear layers, and Jetfire. 
Detailed parameter specifications are provided in the \cref{appendix:exp-set}.
Additionally, we compare the training throughput for DROP with a microbatch size of 2 (1 for Llama-3.1-8B due to memory constraints) and sequence length of 1024 on RTX4090, along with the GPU memory usage of activation contexts across different methods. 
Jetfire's throughput results are not presented as it lacks INT8 dataflow implementation for GLU.
The results are presented in \cref{tab:finetune}.

While the vanilla Block GEMM performs well on most tasks, it suffers from convergence instability in specific cases. 
Jetfire shows significant performance degradation on small models such as Qwen2.5-1.5B and Llama3.2-1B compared to Block GEMM. This decline primarily stems from incompatibility between INT8 dataflow and these models with larger outliers, indicating the importance of accurate Non-Linear operators. 
Our method, however, consistently achieves BF16-comparable accuracy across diverse tasks while maintaining strong robustness.

\begin{figure}[t]
    \centering
    \subfigure[GSM8k/Seed] {
    \includegraphics[width=0.44\linewidth]{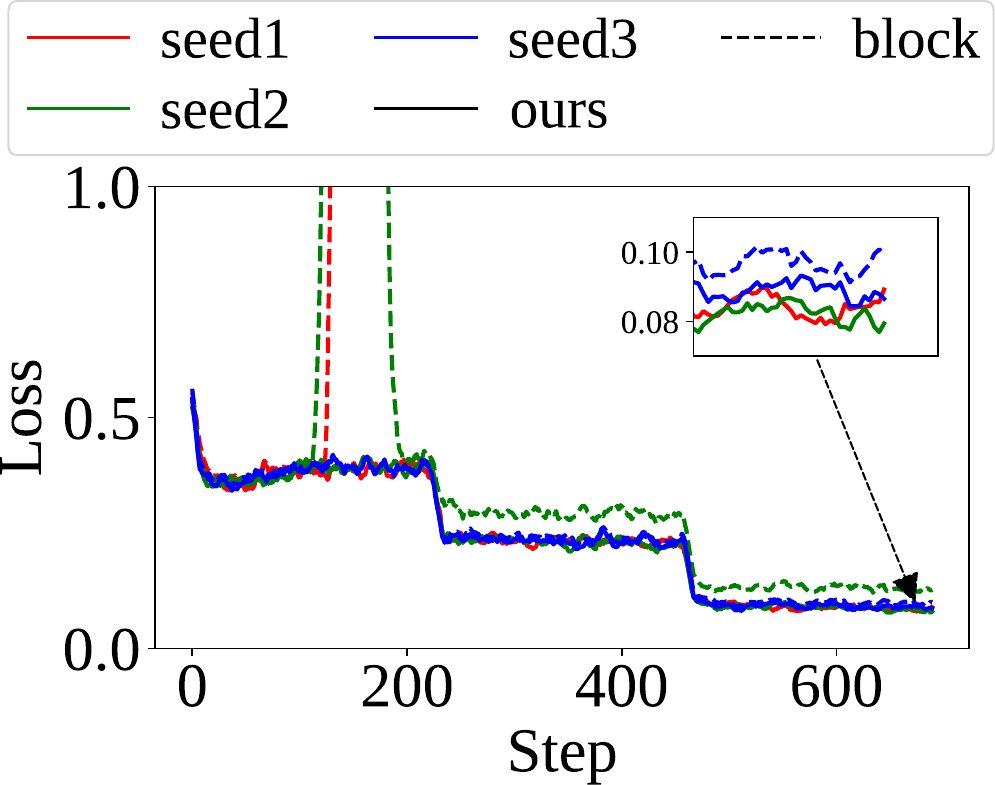}
    \label{fig:finetune-block-gemm-loss}
    }
    \subfigure[GSM8k/FallbackRate] {
    \includegraphics[width=0.43\linewidth]{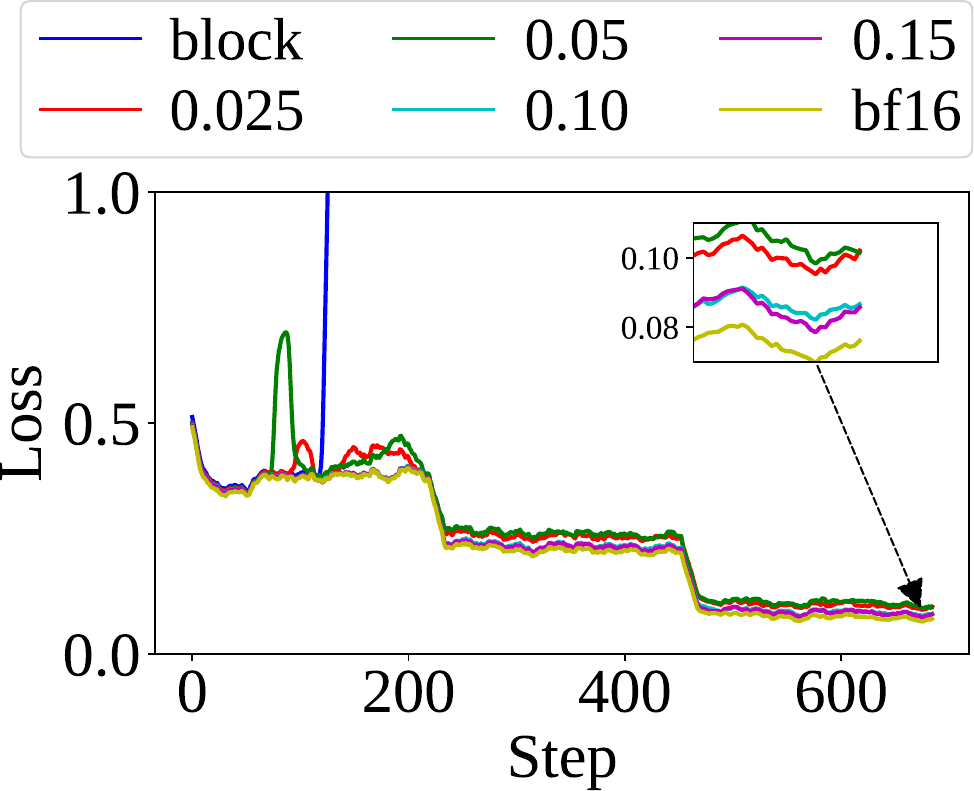}
    \label{fig:finetune-block-gemm-ablation}
    }

    \subfigure[Throughput/FallbackRate]{
        \includegraphics[width=0.9\linewidth]{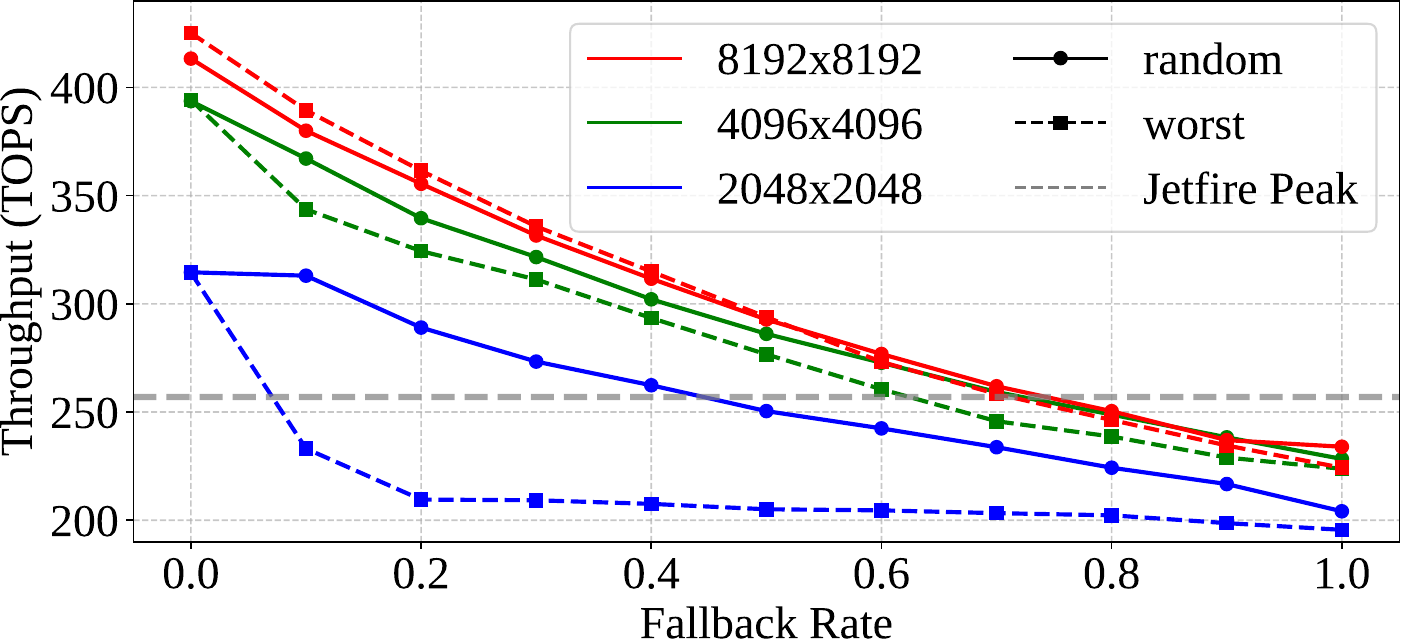}
        \label{fig:kernel-perf}
    }
    \vspace{-.75em}
    \caption{(a) Comparison of loss curves between block and ours under different seeds for qwen2.5-1.5b gsm8k fine-tuning. 
    (b) Comparison of loss curves under different constant fallback rate.
    (c) Fallback GEMM Kernel throughput.}
    
\end{figure}

Block GEMM methods shows significant accuracy degradation in fine-tuning Qwen2.5-1.5B on GSM8K with loss curves diverging across multiple initial seeds. In contrast, our method demonstrates stable performance across various seeds (\cref{fig:finetune-block-gemm-loss}). 
{Our ablation experiments on fallback rate reveal that convergence can be achieved with merely 2.5\% of blocks utilizing fallback, and stable training is maintained with a 10\% fallback rate.}

\subsection{Pretraining}

To validate the effectiveness of our method in pretraining, we trained a Llama-1.5B model on OpenWebText~\cite{Gokaslan2019OpenWeb}. 
The training and validation loss curves are shown in \cref{fig:pretrain-loss}.
Our method's loss curves closely align with BF16, while Jetfire exhibits significant deviations early in training as the introduction of GLU results in wider activation distributions in early stages (Figure~\ref{fig:gpt2-activation-distribution}), making it challenging for block quantization to maintain INT8 data-flow accuracy, while block-quantized GEMM alone can still maintain considerable precision.
Notably, Jetfire exhibits lower training loss but very high validation loss, which we believe is primarily due to information leakage during training, as discussed in the \cref{appendix:information-leakage}.

\subsection{Speedup}
\label{sec:speedup-results}

\begin{table}[th]
\caption{The speedup ratio of a GPT2 transformer layer compared to BF16 under different hidden sizes. We use random inputs with 2 micro batch size and 1024 sequence length. Jet refers to Jetfire.}
\small
\label{tab:linear-speedup}
\vskip 0.15in
\begin{center}
\begin{small}
\begin{tabular}{lcccccc}
\toprule
 & \multicolumn{2}{c}{Forward} & \multicolumn{2}{c}
 {Backward} & \multicolumn{2}{c}{Overall}  \\
 \cmidrule(l){2-3} \cmidrule(l){4-5} \cmidrule(l){6-7}
 Hidden & Jet & Ours & Jet & Ours & Jet & Ours \\
\midrule
1024 & 0.97 & 1.21 & 1.11 & 1.37 & 1.07 & \textbf{1.31}\\
2048 & 1.29 & 1.60 & 1.41 & 1.79 & 1.37 & \textbf{1.73}\\
4096 & 1.32 & 1.56 & 1.46 & 2.16 & 1.42 & \textbf{1.92}\\
\bottomrule
\end{tabular}
\end{small}
\end{center}
\vskip -0.1in
\end{table}

As shown in \cref{tab:finetune}, our method achieves a 1.57x speedup compared to BF16 in end-to-end fine-tuning scenarios. The speedup ratios for different transformer layers and hidden sizes are presented in \cref{tab:linear-speedup}. With larger Block Size, we achieve significant speedup in both forward and backward passes than Jetfire, with the main acceleration at large Hidden Size coming from backward computation.

Fallback leads to varying computational loads across different $C$ blocks, resulting in unstable performance. Blocks with higher computational loads may become the kernel's performance bottleneck, as the GEMM operation must wait for their completion.
We tested the Fallback GEMM Kernel performance under two scenarios: random versus sequential block selection (worst case). For small GEMM (2048), performance loss occurs due to limited C blocks for scheduling optimization. However, since Fallback Blocks typically follow a channel-wise pattern (\cref{fig:kernel-perf}), our method maintains comparable performance to Block GEMM. Detailed results on different GPUs are in Appendix~\ref{appendix:kernel-performance}.
\section{Conclusion}
We propose a novel mixed precision method that increases quantization bits through dynamic block-level fallback quantization and implements an efficient fallback block GEMM to address the limitations of INT8 dynamic range. Our method demonstrates stable accuracy across various tasks and achieves 1.57x end-to-end speedup on RTX4090.

\section*{Impact Statement}
This paper presents work that aims to advance the field of Machine Learning. There are many potential societal consequences of our work, none of which we feel must be specifically highlighted here.

\bibliography{example_paper}
\bibliographystyle{icml2025}

\newpage
\appendix
\onecolumn
\section{Experiment Setting}
\label{appendix:exp-set}

For all fine-tuning tasks, we train for 700 steps using a learning rate of 3e-5 with linear learning rate decay and 100 warmup steps and AdamW with 1e-3 weight decay. Due to the small size of the GSM8K dataset, we use a batch size of 32, while other datasets use a batch size of 128. Our training is conducted on 8 RTX4090 GPUs, with Qwen2.5-3B using a Tensor Parallel (TP) Group Size of 2, and LLaMA 3.1-8B using a TP Group Size of 8. Since our work focuses on computational acceleration, our reported speedup does not include communication overhead.

For pre-training tasks, we use a hidden size of 2048, 20 layers, an intermediate size of 8192, and 16 attention heads. The training uses a learning rate of 1e-3 with linear decay over 2000 warmup steps, and each step processes 1M tokens.
\section{Kernel Performance}
\label{appendix:kernel-performance}
\begin{figure}[h]
    \centering
    \subfigure[3090]{
        \includegraphics[width=0.3\linewidth]{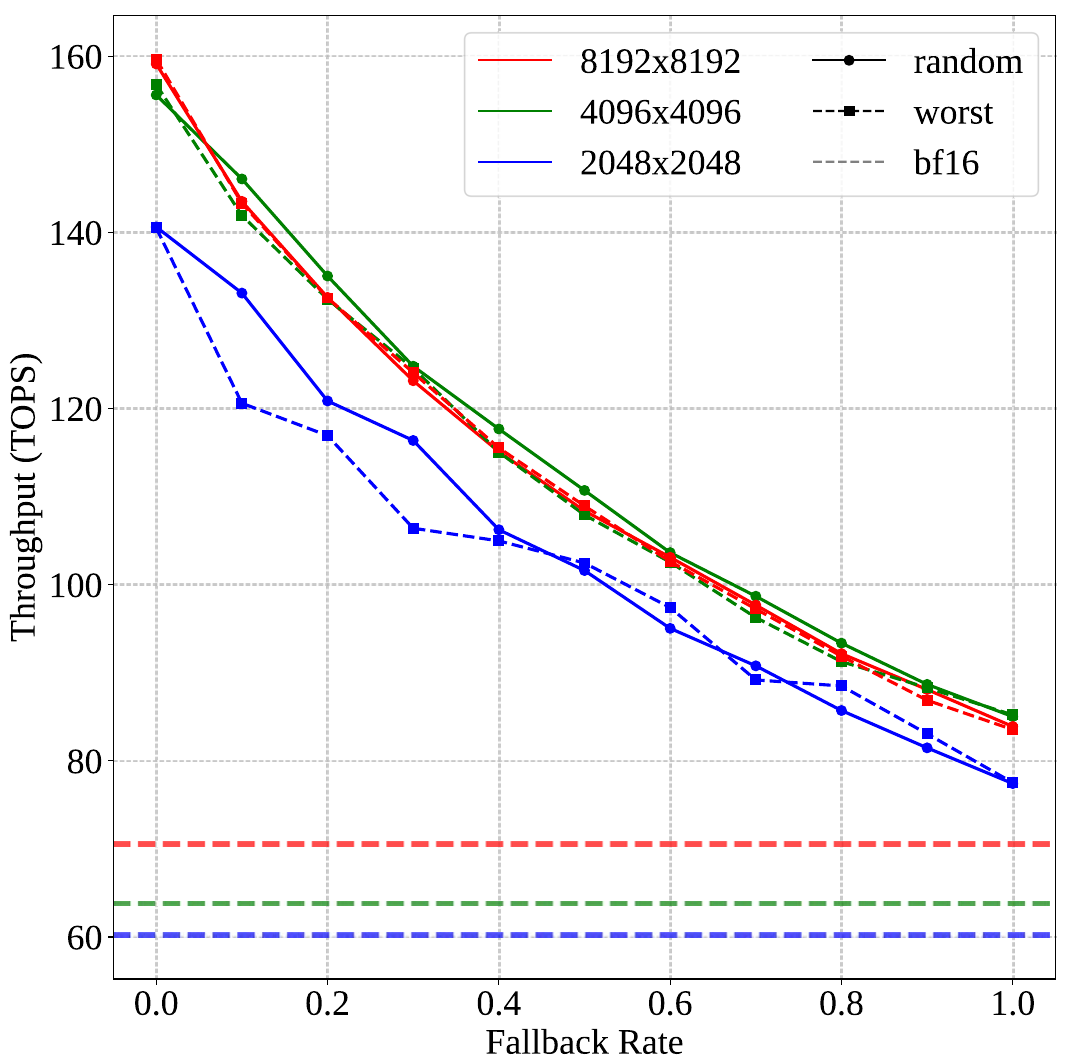}
        \label{fig:3090-kernel-performance}
    }
    \subfigure[L20]{
        \includegraphics[width=0.3\linewidth]{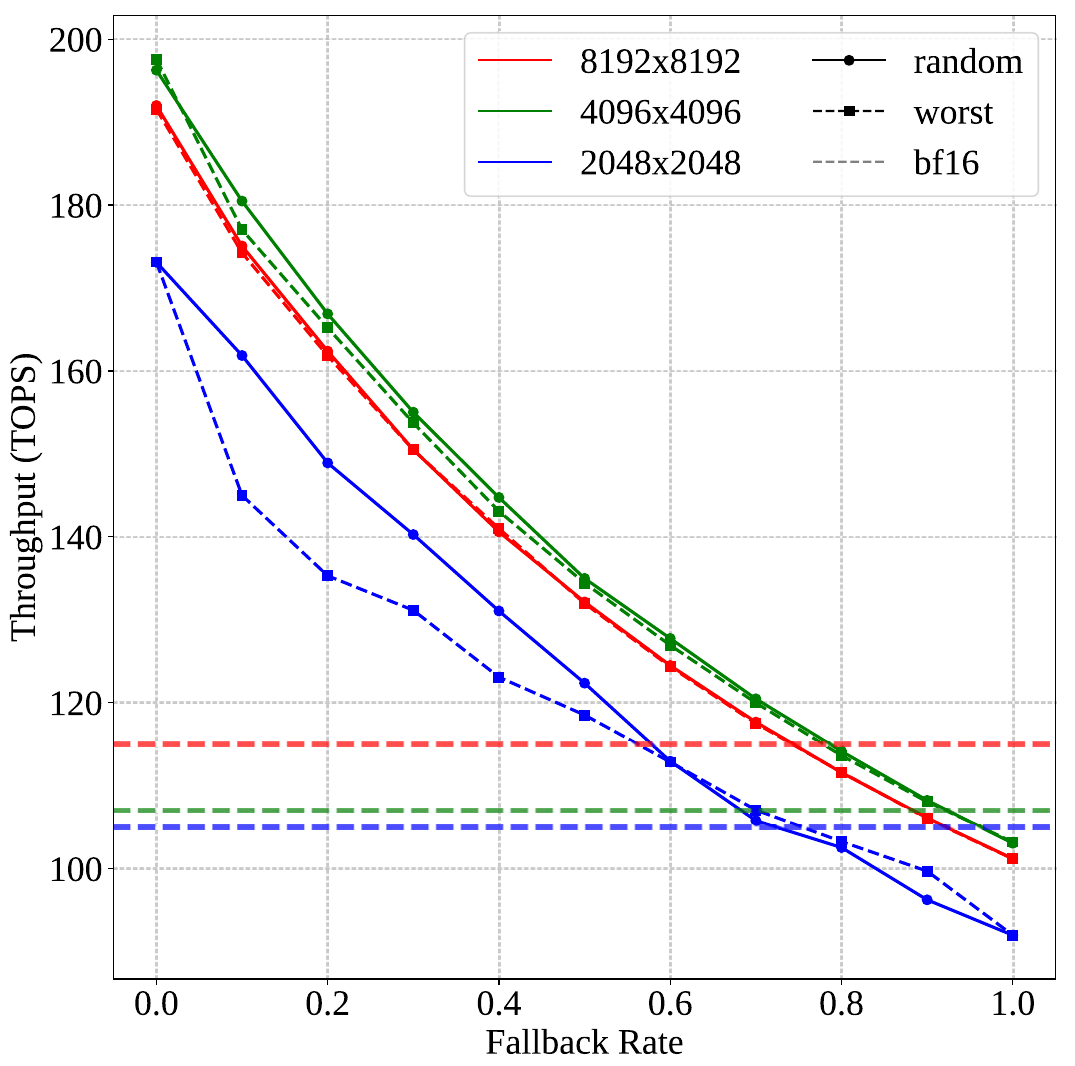}
        \label{fig:L20-kernel-performance}
    }
    \subfigure[A800]{
        \includegraphics[width=0.3\linewidth]{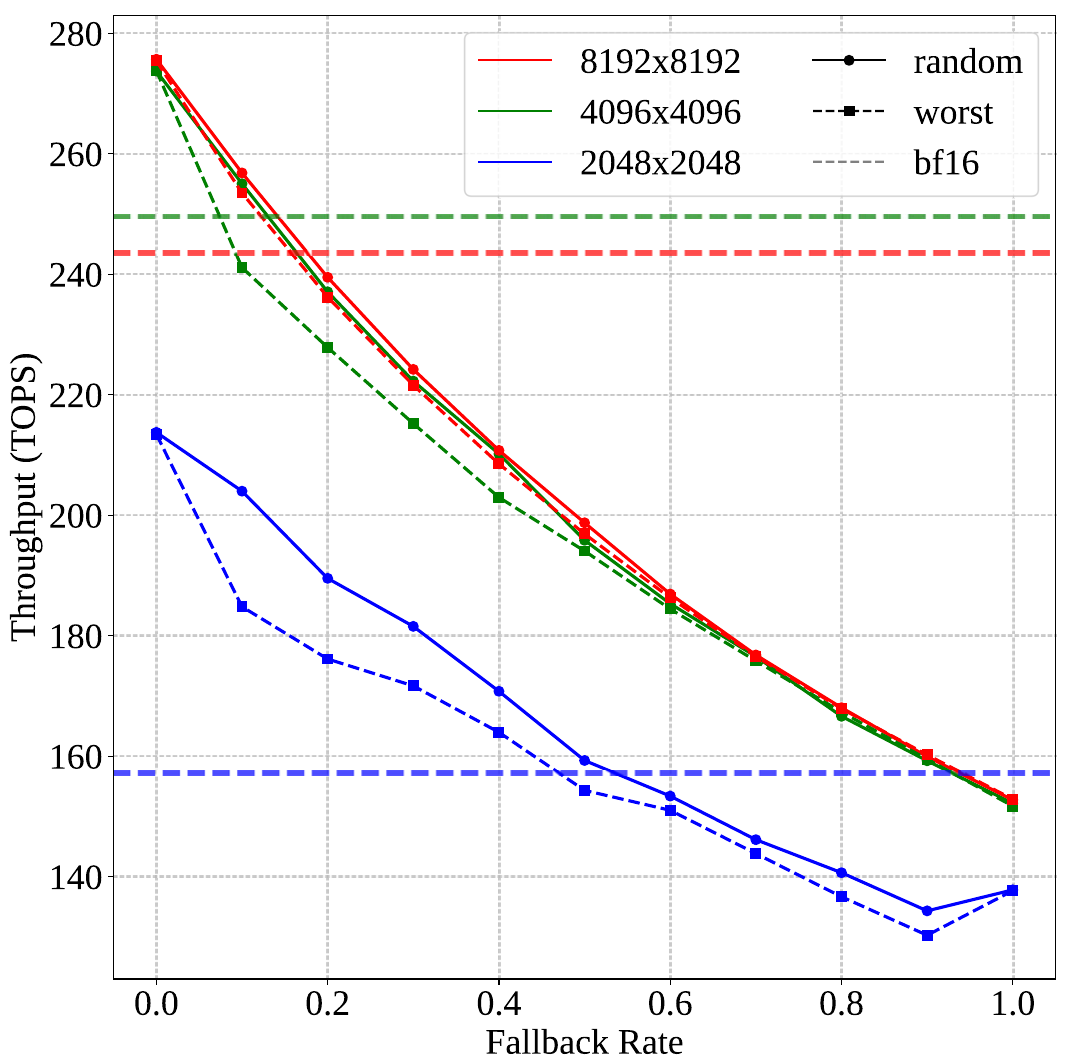}
        \label{fig:A800-kernel-performance}
    }
    \vspace{-.75em}
    
    \caption{Fallback GEMM Kernel throughput on 3090, L20 and A800.}
    \label{fig:other-devices-kernel-performance}
\end{figure}

We also tested the performance of the Fallback GEMM Kernel on three other types of GPUs: L20, 3090 and A800. 
Simliar to \cref{fig:kernel-perf}, \cref{fig:other-devices-kernel-performance} shows the scenarios of the random fallback A Tiles and the worst sequential fallback A Tiles.

Our Fallback GEMM Kernel achieves up to 2.47x speedup on the 3090 and 1.85x speedup on the L20, which can be attributed to the four times theoretical peak flops of INT8 as BF16. 
While on A800 the corresponding ratio is just twice and CUDA Cores are insufficient for dequantization, our kernel gains less on speed. However, we can still benefit from reduced memory consumption. Besides, we can inference with other more INT8-friendly devices.

\section{Different Block Sizes}
\label{appendix:different-block-sizes}

The quantization block size refers to the granularity of quantization, while the GEMM tile size represents the block size of GEMM operators on GPUs.

It is not necessary to restrict the quantization block size and the GEMM tile size to be identical. Typically, the GEMM tile size optimized for specific device is not greater than our selected quantization block size $128 \times 128 \times 128$. For example, on 4090, the tile size is $128 \times 128 \times 128$
, and on L20, the tile size is $64 \times 64 \times 64$. In this context, a tile is one sub-block of a quantization block, and GEMM operations are performed on these sub-blocks, using the same scale factor across them.

\section{Delay Threshold}
\label{appendix:delay-threshold}
{
Training is inherently dynamic, and a fixed 
threshold may result in either excessive or insufficient fallback rates, impacting either performance or accuracy. 
While selecting TopK AbsMax from the current input as the threshold requires multiple operations and reduction across the entire tensor. 
To efficiently update the threshold dynamically, we draw inspiration from the Delay-Scaling approach used in quantization~\cite{micikevicius2022fp8}: the current threshold is determined by the Fallback Rate from previous steps.

Specifically, we implement layer-specific Fallback Thresholds for each Linear layer. To maintain a balance between accuracy and performance, we set both lower and upper bounds for the Fallback Rate $[r_{min}, r_{max}]$. With a global Adjustment Factor $\alpha$, the thresholds are dynamically adjusted after each training iteration: 
for any Linear layer, the threshold is decreased by dividing by $\alpha$ when its Fallback Rate falls below $r_{min}$, and increased by multiplying by $\alpha$ when it exceeds $r_{max}$. This delay threshold procedure is formally described in Algorithm \ref{alg:delay-threshold}.
}

\begin{algorithm}[thp]
   \caption{Delay Threshold}
   \label{alg:delay-threshold}
\begin{algorithmic}[1]
   \STATE {\bfseries Input:}
   \STATE Model $\mathcal{M}$
   \STATE Training data batches $\mathcal{D}$
   \STATE Adjustment factor $\alpha > 1$
   \STATE Target fallback rate range $[r_{min}, r_{max}]$
   \STATE {\bfseries Initialize:}
   \FOR{each linear layer $L$ in $\mathcal{M}$}
       \STATE $L.threshold \leftarrow 1$
   \ENDFOR
   \STATE {\bfseries Training:}
   \FOR{batch $B$ in $\mathcal{D}$}
       \STATE Update model $\mathcal{M}$ with batch $B$
       \FOR{each linear layer $L$ in $\mathcal{M}$}
           \IF{$L.fallback\_rate < r_{min}$}
               \STATE $L.threshold \leftarrow L.threshold/\alpha$
           \ELSIF{$L.fallback\_rate > r_{max}$}
               \STATE $L.threshold \leftarrow L.threshold \times \alpha$
           \ENDIF
       \ENDFOR
   \ENDFOR
\end{algorithmic}
\end{algorithm}
\section{Information Leakage}
\label{appendix:information-leakage}

As we have discussed in \cref{section:exp}, Jetfire shows significant gain on pretrain training loss 
but does poorly on evaluation. 
This stems from fine-grain quantization with block size $32\times 32$, 
which is small enough for model to utilize its AbsMax information 
to receive information from next token, making training loss lower.

We can verify this hypothesis through different evaluation methods. 
The first method is to test the model using BF16 precision, which will not lead to information leakage. 
Another method is to incorporate quantization during evaluation, while still inputting the entire text in one evaluation iteration and calculating the loss. 
Finally, we can incorporate quantization during evaluation, but predict tokens one at a time without access to future tokens, which prevents information leakage.

\begin{table}[th]
\caption{Validation perplexities across different validation settings on the 30B tokens pretrain checkpoint. 
We disable fallback quantization on checkpoint of our method for fair comparison. }
\small
\label{tab:pretrain-valid-loss}
\vskip 0.15in
\begin{center}
\begin{tabular}{lcccc}
\toprule
 Method & BF16 & Jetfire & Block-Gemm & Ours \\
\midrule
 BF16             & 14.17 & 24.14 & 14.73 & \textbf{14.19} \\
Quant             & -    & \textbf{12.24} & \textbf{14.10} & 14.25 \\
Quant(no leakage) & -    & 26.08 & 14.79 & 14.26 \\
\bottomrule
\end{tabular}
\end{center}
\vskip -0.1in
\end{table}

In \cref{tab:pretrain-valid-loss}, Jetifre shows significant degration 
when using BF16 or quantized inference without information leakage, and even Block GEMM shows lower PPL when using quantized inference.
Our method, however, performs best under BF16 precision, which means that the information learned by the model is not related to quantization. This is because the fallback mechanism renders ineffective the information leakage caused by AbsMax.


\end{document}